\documentclass[3p]{elsarticle}

\makeatletter
\def\ps@pprintTitle{%
  \let\@oddhead\@empty
  \let\@evenhead\@empty
  \def\@oddfoot{\reset@font\hfil\thepage\hfil}
  \let\@evenfoot\@oddfoot
}
\makeatother

\usepackage{hyperref}
\usepackage[labelfont=bf,font=normalsize,figurename=Fig.,justification=raggedright,]{caption}
\usepackage{subcaption}

\usepackage{array}
\usepackage{multirow}

\newcommand{\specialcell}[2][c]{%
  \begin{tabular}[#1]{@{}l@{}}#2\end{tabular}}

\begin{document}
\begin{frontmatter}
    \title{IGUANe: a 3D generalizable CycleGAN for multicenter harmonization of brain MR images}
    \author[1]{Vincent Roca\corref{cor1}}
    \author[1,2,3]{Grégory Kuchcinski}
    \author[1,2,3]{Jean-Pierre Pruvo}
    \author[1]{Dorian Manouvriez}
    \author[1,2,4]{Renaud Lopes}
    \author{the Australian Imaging Biomarkers and Lifestyle flagship study of ageing\fnref{fnAibl}}
    \author{the Alzheimer’s Disease Neuroimage Initiative\fnref{fnAdni}}

    \cortext[cor1]{Corresponding author at: Lille, 59037 France. E-mail address: vincentroca9@outlook.fr}

    \address[1]{Univ. Lille, CNRS, Inserm, CHU Lille, Institut Pasteur de Lille, US 41 - UAR 2014 - PLBS, F-59000 Lille, France}
    \address[2]{Univ. Lille, Inserm, CHU Lille, U1172 - LilNCog - Lille Neuroscience \& Cognition,
F-59000 Lille, France}
    \address[3]{CHU Lille, Département de Neuroradiologie, F-59000 Lille, France}
    \address[4]{CHU Lille, Département de Médecine Nucléaire, F-59000 Lille, France}

    \fntext[fnAibl]{Data used in the preparation of this article was obtained in part from the Australian Imaging Biomarkers and Lifestyle flagship study of ageing (AIBL) funded by the Commonwealth Scientific and Industrial Research Organisation (CSIRO) which was made available at the ADNI database (www.loni.usc.edu/ADNI). The AIBL researchers contributed data but did not participate in analysis or writing of this report. AIBL researchers are listed at \url{www.aibl.csiro.au}.}

    \fntext[fnAdni]{Data used in preparation of this article were obtained in part from the Alzheimer’s Disease Neuroimaging Initiative (ADNI) database (adni.loni.usc.edu). As such, the investigators within the ADNI contributed to the design and implementation of ADNI and/or provided data but did not participate in analysis or writing of this report. A complete listing of ADNI investigators can be found at: \url{http://adni.loni.usc.edu/wp-content/uploads/how_to_apply/ADNI_Acknowledgement_List.pdf}}

    \begin{abstract}
        In MRI studies, the aggregation of imaging data from multiple acquisition sites enhances sample size but may introduce site-related variabilities that hinder consistency in subsequent analyses. Deep learning methods for image translation have emerged as a solution for harmonizing MR images across sites. In this study, we introduce IGUANe (Image Generation with Unified Adversarial Networks), an original 3D model that leverages the strengths of domain translation and straightforward application of style transfer methods for multicenter brain MR image harmonization. IGUANe extends CycleGAN by integrating an arbitrary number of domains for training through a many-to-one architecture. The framework based on domain pairs enables the implementation of sampling strategies that prevent confusion between site-related and biological variabilities. During inference, the model can be applied to any image, even from an unknown acquisition site, making it a universal generator for harmonization. Trained on a dataset comprising T1-weighted images from 11 different scanners, IGUANe was evaluated on data from unseen sites. The assessments included the transformation of MR images with traveling subjects, the preservation of pairwise distances between MR images within domains, the evolution of volumetric patterns related to age and Alzheimer’s disease (AD), and the performance in age regression and patient classification tasks. Comparisons with other harmonization and normalization methods suggest that IGUANe better preserves individual information in MR images and is more suitable for maintaining and reinforcing variabilities related to age and AD. Future studies may further assess IGUANe in other multicenter contexts, either using the same model or retraining it for applications to different image modalities. Codes and the trained IGUANe model are available at \url{https://github.com/RocaVincent/iguane_harmonization.git}.
    \end{abstract}

\end{frontmatter}

\section{Introduction}
In the last few years, MRI data from multiple acquisition sites have become more accessible, leading to an increase in sample sizes and, consequently, enhanced statistical power in brain MRI studies. However, heterogeneities related to technical factors, including manufacturer, field strength, sequence design, and RF coil, have been reported \citep{Chen2014,Han2006}. These differences can impact statistical analyses, even when efforts have been made to standardize acquisition protocols \citep{Kruggel2010,Shinohara2017}. Technical variabilities in medical images can also limit the generalizability of predictive models based on machine learning \citep{Bashyam2022,Zech2018}.

Various retrospective harmonization methods have been proposed to face this problem. Three broad categories exist: (i) statistical and machine learning methods harmonizing image features, (ii) normalization to a standard intensity scale, and (iii) deep learning models processing MR data at the voxel level. The first category primarily consists of methods derived from ComBat \citep{Chen2022,Fortin2018,Pomponio2020}, which models technical variabilities as additive and multiplicative factors for each input feature and each site \citep{Fortin2017}. Efficiency was demonstrated for specific applications but these methods require different harmonization procedures for each set of features and are not well-suited for processing whole brain images. In the second category, statistics from each image are computed and used to normalize them independently. The simplicity of these methods make them widely used in MR multicenter studies, such as min-max \citep{Naser2020,Li2020,Gautherot2021} and z-score \citep{Welander2018,Dinsdale2021} normalizations. The approaches in the third category are based on deep neural networks, enabling the modeling of more complex patterns. Some of these methods are tailored for specific prediction tasks \citep{Dinsdale2021,Guan2021,Wang2022}, which necessitates a separate training process for each application and restricts their use to predictive contexts. 

Many recent works have therefore focused on developing image-to-image translation (IIT) models. Some of them are based on supervised learning, proving efficient with small datasets but requiring traveling subjects \citep{Dewey2019,Tian2022,Torbati2023}. In contrast, many recently proposed deep learning models operate in an unsupervised manner, enabling harmonization without the need of traveling subjects. However, certain limitations persist, hindering widespread adoption by end-users: potential loss of biological information, insufficient validation on external datasets, important computational resource requirements, the lack of readily usable code for method reuse, and the independent processing of 2D slices from the same volume \citep{Cohen2018,Hu2023}.

To address these limitations, we propose IGUANe (Image Generation with Unified Adversarial Networks), an unsupervised IIT model designed for inter-site harmonization of structural brain MR images. Based on a many-to-one strategy built upon the CycleGAN framework, it enables the harmonization of any MR image after training while avoiding the loss of biological information. Its 3D architecture is designed to process whole brain volumes. IGUANe is publicly available and usable without important computational requirements.

To evaluate the effectiveness of our tool, we conducted experiments on T1-weighted (T1w) brain MR images using a multisite training set and test sets from various external studies. We validated and compared our approach to other methods by examining image similarity metrics, the preservation and enhancement of biological patterns, and accuracy in prediction tasks. The results showcase the model’s ability to efficiently harmonize MR images from unseen sites.

\section{Related works}

In this section, we focus on unsupervised IIT employed for harmonization of structural brain MR images.

\subsection{Unsupervised frameworks}
The first category of unsupervised IIT is domain translation, CycleGAN \citep{Zhu2017} being one of the most renowned models. Studies have demonstrated their efficacy in harmonizing brain MR images \citep{Chen2021,Enriquez2021,Gebre2023,Nguyen2018,Palladino2020,Roca2023}. However, CycleGAN requires training for every pair of sites and may not fully leverage complementary information from all sites. Other domain translation models addressing this issue include derivations of StarGan v2 \citep{Bashyam2022,Liu2024}, a many-to-one CycleGAN \citep{Gao2019} and a style-blind auto-encoder \citep{Fatania2022}.

The second category of unsupervised IIT is style transfer, focusing on image-level style rather than domain-level characteristics \citep{Gatys2016}. For instance, the model of \citet{Zuo2022} is trained with assumptions about MRI volumes: slices at different locations share the same contrast but have different anatomic information. Similar assumptions guide other methods, supplemented by self-supervised learning through pairing different MRI contrasts \citep{Zuo2021} or creating pairings with image processing functions \citep{Cackowski2023}. The simplified version of StarGAN v2 proposed by \citet{Liu2023} does not use domain information.

In both categories of unsupervised IIT, with the exception of \citet{Roca2023}, almost all existing models have been designed for processing slices or low-volume patches. The resultant loss of spatial and contextual information in these approaches is often either unjustified or dictated by technical limitations \citep{Dewey2020,Liu2023,Nguyen2018, Palladino2020,Zuo2021}.

Compared to style transfer models, domain translation models aim to learn higher-level correspondences between sets of images. We therefore combined and extended two existing domain translation works \citep{Gao2019,Roca2023} to set up a 3D model able to harmonize MR images from unseen sites while preventing the loss of biological information (section \ref{sec:iguane}).

\subsection{Evaluation metrics}
The use of image similarity metrics on datasets with traveling subjects is a prevalent practice in assessing IIT models for harmonization \citep{Cackowski2023,Gao2019,Liu2023,Liu2024,Nguyen2018,Robinson2020,Roca2023,Zuo2021,Zuo2022}. These datasets, providing ground truth for harmonization, are well-suited for evaluation. However, these validations are often constrained to a small number of subjects scanned at a few sites. Consequently, they are supplemented by assessments of similarities between sites before and after harmonization. This includes site prediction \citep{Cackowski2023,Nguyen2018}, visualization of MRI data decomposed in two-dimensional spaces \citep{Cackowski2023,Liu2023,Nguyen2018,Roca2023}, quantification of differences in feature distributions across sites \citep{Liu2024,Roca2023} and comparison of intensity histograms \citep{Cackowski2023,Roca2023}. However, these evaluations don’t account for biological patterns and are applicable only to MR images that can be clustered into domains (e.g. sites). Predictive models have therefore been set up to assess the preservation of relevant information in the original images, for example with brain age prediction \citep{Bashyam2022,Liu2023,Robinson2020,Roca2023}, classification of sex \citep{Nguyen2018,Robinson2020}, pathological status \citep{Cackowski2023,Gao2019,Nguyen2018} and brain tissue segmentation \citep{Chen2021}. Furthermore, automatic segmentation tools have been used to study the impact of harmonization for the detection of volumetric patterns related to age \citep{Roca2023} and Alzheimer’s disease (AD) \citep{Liu2023} in MR images.

Most of these validations have relied on MR images from sites included in the training of harmonization models. Experiments assessing generalization to unseen sites have been largely limited to visual assessments and image similarity metrics with traveling subjects \citep{Cackowski2023,Gao2019,Liu2023}. In this study, we applied our proposed harmonization model to unseen sites and analyzed its impact on age and pathological information (section \ref{sec:experiments}).

\section{Materials and Methods}

\subsection{MRI datasets}
We used T1w brain MR images from public databases for this study, limiting inclusion to participants aged between 18 and 80 years. Metadata enabling the identification of the used images are available in our online repository. The main characteristics of each dataset are given below. The list of image resolutions before preprocessing are listed in \ref{appendix_resolutions}.

\paragraph{Training dataset} To train our harmonization (section \ref{sec:iguane_implementation}) and brain age (section \ref{sec:meth_prediction}) models, we included MR images from eight studies: SALD \citep{Wei2018}, IXI\footnotemark[3], OASIS-3 \citep{LaMontagne2019}, NKI-RS \citep{Nooner2012}, NMorphCH\footnotemark[4], AIBL \citep{Ellis2009}, HCP Young Adult\footnotemark[5] and the International Consortium for Brain Mapping\footnotemark[6] (ICBM). All participants were healthy controls. MR images were acquired using eleven different machines, referred to as domains in our harmonization model (section 3.2). Each participant was exclusively present in a single domain. Demographic and scanner information is given in Table \ref{tab:training_dataset}.

\begin{table}
    \centering
    \caption{\textbf{Characteristics of the healthy participants and the scanners in the Training dataset.}}
    \begin{tabular}{>{\raggedright}p{0.10\linewidth}>{\raggedright}p{0.22\linewidth}>{\raggedright}p{0.09\linewidth}>{\raggedright}p{0.15\linewidth}>{\raggedright}p{0.12\linewidth}>{\raggedright\arraybackslash}p{0.1\linewidth}}
        \hline
        \textbf{Study}&\textbf{Scanner model}&\textbf{Field strength, Tesla}&\textbf{Nb of MR images / nb of participants}&\textbf{Age, years}$^\#$&\textbf{Males, \%}\\
        \hline
        SALD&Siemens Magnetom TrioTim&3&494/494&$45.18 \pm 17.44$&38\\
        IXI&Philips Intera&1.5&305/305&$47.31 \pm 16.52$&48\\
        IXI&Philips Intera&3&176/176&$50.20 \pm 15.46$&43\\
        OASIS-3&Siemens Magnetom TrioTim&3&857/350&$67.03 \pm 8.27$&33\\
        OASIS-3&Siemens Biograph mMR PET-MR&3&412/311&$68.10 \pm 7.51$&46\\
        NKI-RS&Siemens Magnetom TrioTim&3&249/247&$29.97 \pm 8.20$&40\\
        NMorphCH&Siemens Magnetom TrioTim&3&141/44&$31.37 \pm 8.42$&53\\
        AIBL&Siemens Magnetom TrioTim&3&489/280&$71.93 \pm 4.90$&46\\
        HCP&Siemens Connectome Skyra&3&402/402&$28.90 \pm 3.71$&39\\
        ICBM&Siemens Sonata&1.5&677/135&$43.97 \pm 15.21$&48\\
        ICBM&Philips ACS III&1.5&145/145&$25.10 \pm 4.93$&56\\
        \hline
        \multicolumn{6}{@{}l}{$^\#$ Age is expressed as mean $\pm$ standard deviation.}\\
    \end{tabular}
    \label{tab:training_dataset}
\end{table}

\paragraph{Traveling subject dataset} We used the SRPBS traveling subject MRI dataset \citep{Tanaka2021} to study the evolution of similarities between MR images with participants scanned with different MR models (section \ref{sec:meth_travSubs}). This dataset includes brain MR images from 9 healthy male participants (age range 24-32 years). In this study, we used a total of 97 images acquired with 11 different machines (GE, Siemens and Philips manufacturers).

\paragraph{Generalization dataset} To visualize image harmonizations (section \ref{sec:visua_harmonization}) and to study the evolution of aging patterns (sections \ref{sec:meth_corrGMage} and \ref{sec:meth_prediction}), we included MR images from five studies that were unseen during the training phase: the Alzheimer’s Disease Neuroimaging Initiative\footnotemark[7] (ADNI), the Mind Clinical Imaging Consortium (MCIC) database \citep{Gollub2013}, PPMI\footnotemark[8] (data openly available), COBRE \citep{Aine2017} and ABIDE\footnotemark[9]. All participants were healthy controls. Demographic and scanner information is given in Table \ref{tab:generalization_dataset}.  

\begin{table}
    \centering
    \caption{\textbf{Characteristics of the healthy participants and the scanners in the Generalization dataset.}}
    \begin{tabular}{>{\raggedright}p{0.08\linewidth}>{\raggedright}p{0.16\linewidth}>{\raggedright}p{0.14\linewidth}>{\raggedright}p{0.18\linewidth}>{\raggedright}p{0.13\linewidth}>{\raggedright\arraybackslash}p{0.1\linewidth}}
        \hline
        \textbf{Study}&\textbf{Manufacturer}$^\#$&\textbf{Field strength, Tesla}$^\#$&\textbf{Nb of MR images / nb of participants}&\textbf{Age, years}$^\dag$&\textbf{Males, \%}\\
        \hline
        ADNI&\specialcell[t]{GE (115);\\Philips (109);\\Siemens (104)}&\specialcell[t]{3 (199);\\1.5 (129)}&328/216&$73.62 \pm 4.37$&45\\
        MCIC&Siemens&\specialcell[t]{1.5 (192);\\3 (52)}&244/89&$34.49 \pm 11.93$&67\\
        PPMI&\specialcell[t]{Siemens (202);\\Philips (39)}&\specialcell[t]{3 (215);\\1.5 (24);\\unknown (2)}&241/141&$60.49 \pm 10.65$&64\\
        COBRE&Siemens&3&227/91&$38.57 \pm 11.55$&73\\
        ABIDE&\specialcell[t]{Siemens (105);\\Philips (34)}&3&139/139&$27.40 \pm 6.83$&88\\
        \hline
        \multicolumn{6}{@{}l}{$^\#$ The number of MR images is indicated in brackets if there are several options.}\\
        \multicolumn{6}{@{}l}{$^\dag$ Age is expressed as mean $\pm$ standard deviation.}\\
    \end{tabular}
    \label{tab:generalization_dataset}
\end{table}

\paragraph{Clinical dataset} We used data from ADNI, AIBL and MIRIAD \citep{Malone2013} to investigate the impact of harmonization on patterns related to AD (sections \ref{sec:meth_hippoVols} and \ref{sec:meth_prediction}). We specifically selected participants diagnosed as cognitively normal (CN) or with AD. To conduct a CN/AD classification (section \ref{sec:meth_prediction}), we divided the collection in four sets: \textit{AD\_train} for training, \textit{AD\_test} for testing, \textit{AD\_GE} for generalization to MR images from GE manufacturer (no GE MR image is in the AD\_train, AD\_test nor Training datasets) and \textit{MIRIAD} for generalization to MR images from another study. Importantly, none of the participants in the AD\_train dataset were included in the other three datasets. Additionally, no subject or scanner was present during the training of the harmonization model. Demographic and scanner information is detailed in Table \ref{tab:clinical_dataset}.

\begin{table}
    \centering
    \caption{\textbf{Characteristics of the participants and the scanners in the Clinical dataset.}}
    \begin{tabular}{>{\raggedright}p{0.08\linewidth}>{\raggedright}p{0.14\linewidth}>{\raggedright}p{0.08\linewidth}>{\raggedright}p{0.13\linewidth}>{\raggedright}p{0.08\linewidth}>{\raggedright}p{0.06\linewidth}>{\raggedright}p{0.06\linewidth}>{\raggedright}p{0.03\linewidth}>{\raggedright\arraybackslash}p{0.03\linewidth}}
        \hline
         \textbf{Dataset}&\textbf{Manufacturer}$^\#$&\textbf{Field strength, Tesla}$^\#$&\textbf{Nb of MR images / nb of participants}&\textbf{CN/AD, \%}$^\dag$&\multicolumn{2}{l}{\textbf{Age, years}$^\ddag$}&\multicolumn{2}{l}{\textbf{Males, \%}}\\
         &&&&&\textbf{CN}&\textbf{AD}&\textbf{CN}&\textbf{AD}\\
         \hline
         AD\_train&\specialcell[t]{Siemens (1286);\\Philips (489)}&\specialcell[t]{3 (1132);\\1.5 (643)}&1775/546&50/50&$72.44 \pm 5.24$&$72.02 \pm 5.83$&44&53\\
         AD\_test&\specialcell[t]{Siemens (518)\\Philips (169)}&\specialcell[t]{3 (317);\\1.5 (270)}&687/237&50/50&$71.97 \pm 5.05$&$71.62 \pm 6.63$&49&49\\
         AD\_GE&GE&\specialcell[t]{3 (642);\\1.5 (653)}&1295/275&50/50&$73.12 \pm 4.57$&$71.69 \pm 5.79$&46&53\\
         MIRIAD&GE&1.5&652/64&33/67&$69.86 \pm 6.94$&$69.56 \pm 6.86$&46&41\\
         \hline
         \multicolumn{9}{@{}l}{$^\#$ The number of MR images is indicated in brackets if there are several options.}\\
         \multicolumn{9}{@{}l}{$^\dag$ CN: cognitively normal; AD: Alzheimer’s disease}\\
         \multicolumn{9}{@{}l}{$^\ddag$ Age is expressed as mean $\pm$ standard deviation.}\\
    \end{tabular}
    \label{tab:clinical_dataset}
\end{table}

\subsection{IGUANe model}
\label{sec:iguane}

In this section, we present the main characteristics of the IGUANe model. Additional information is given in \ref{appendix_iguane}.

\subsubsection{MRI preprocessing}
\label{sec:preproc}
To eliminate the most trivial technical variabilities in the data and facilitate harmonization, we preprocessed the MR images as follows: (i) skull-stripping with HD-BET \citep{Isensee2019}, (ii) bias correction with N4ITK \citep{Tustison2010}, (iii) linear registration to 1 mm$^3$ MNI space with FSL-FLIRT \citep{Jenkinson2002} (six degrees of freedom), (iv) cropping to 160 x 192 x 160 voxels, and (v) division of intensities by the median brain intensity. This last step standardizes the median intensity inside the brain while keeping the background at 0. We found it to be more robust to outliers than the classical rescaling approach.

\subsubsection{Universal generator}
\label{sec:universal_generator}
The IGUANe framework is based on a many-to-one strategy in which a generator, $GenFwd$, is trained to translate images from $N$ source sites ($Site_1$, $Site_2$, …, $Site_N$) to a reference site, $SiteRef$. To apply cycle-consistency constraints \citep{Zhu2017}, $N$ backward generators ($GenBwd_1$, $GenBwd_2$, …, $GenBwd_N$) are set up to translate images from $SiteRef$ to each source site. For adversarial training, $N$ forward discriminators ($DiscFwd_1$, $DiscFwd_2$, …, $DiscFwd_N$) learn to distinguish real $SiteRef$ images from harmonized images generated by $GenFwd$ from each source site. Similarly, $N$ backward discriminators ($DiscBwd_1$, $DiscBwd_2$, …, $DiscBwd_N$) learn to distinguish real images from each source site from images harmonized with the backward generators.

\begin{figure}
    \centering
    \begin{subfigure}{0.4\linewidth}
        \centering
        \includegraphics[scale=0.5]{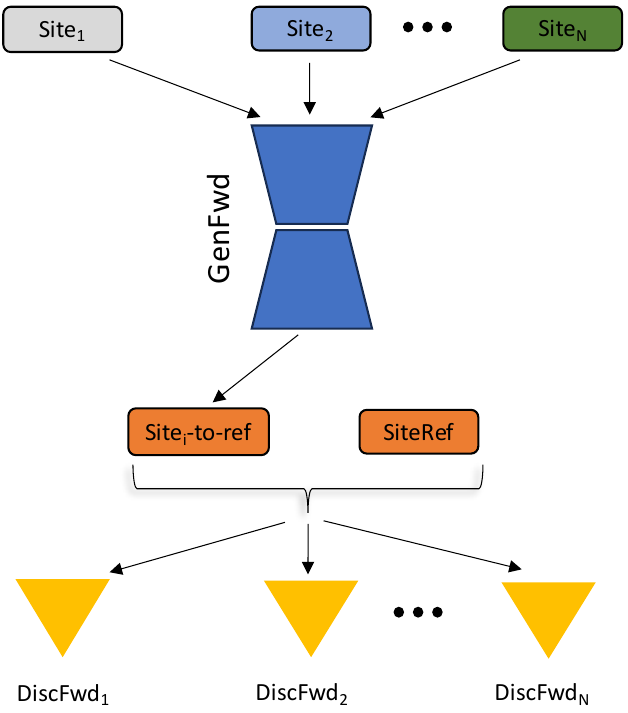}
        \caption{Harmonization towards $SiteRef$}
    \end{subfigure}
    \begin{subfigure}{0.5\linewidth}
        \centering
        \includegraphics[scale=0.5]{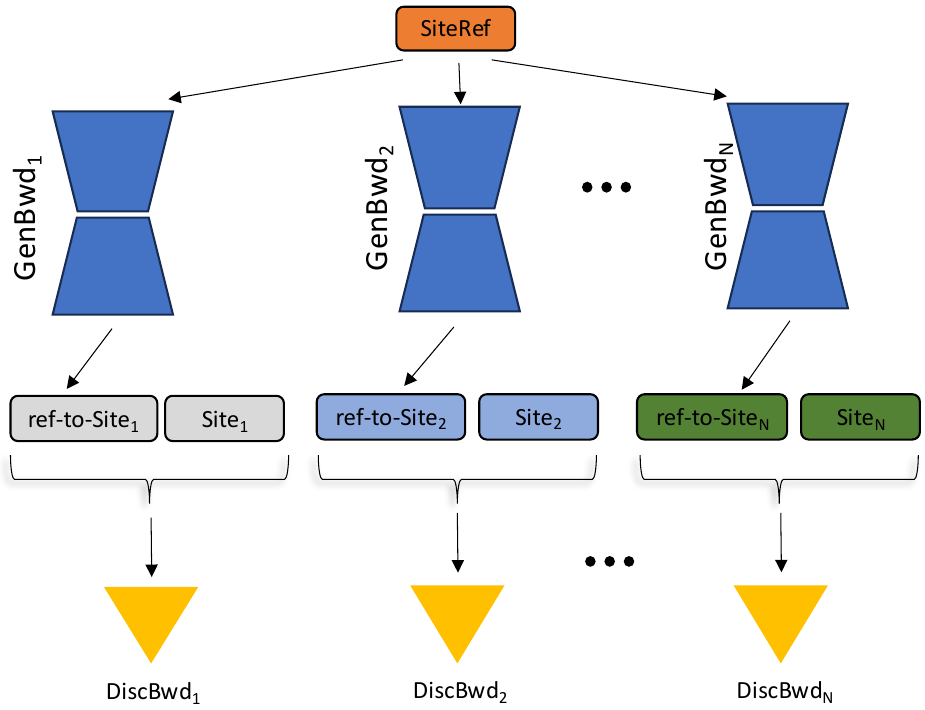}
        \caption{Harmonization towards the source sites.}
    \end{subfigure}
    \caption{\textbf{Representation of the modules in the IGUANe framework.}}
    \label{fig:iguane_modules}
\end{figure}

The modules of the framework are illustrated in Fig. \ref{fig:iguane_modules}. During inference, $GenFwd$ is used to harmonize MR images. The assumption is that a sufficiently diverse training set in terms of acquisition sites and biological characteristics will enable the harmonization of any MR image, including those with an unknown acquisition site.

This many-to-one configuration is inspired by \citet{Gao2019} who used a similar strategy to harmonize T2-FLAIR slices. The main difference, beyond the training procedure, network architectures and implementation tips (sections \ref{sec:training_procedure}, \ref{sec:network_architectures} and \ref{sec:iguane_implementation}, respectively), lies in the $N$ forward discriminators, whereas \citeauthor{Gao2019} included a single one in their model. The motivation behind this extension is to enable a sampling strategy specific to each source site (section \ref{sec:iguane_implementation}), aiming to avoid confusion between technical and biological variabilities. We analyzed the impact of this extension in ablation studies (section \ref{sec:meth_ablationStudies}).

\subsubsection{Training procedure}
\label{sec:training_procedure}
\begin{figure}
    \centering
    \begin{subfigure}{\linewidth}
        \centering
        \includegraphics[scale=0.4]{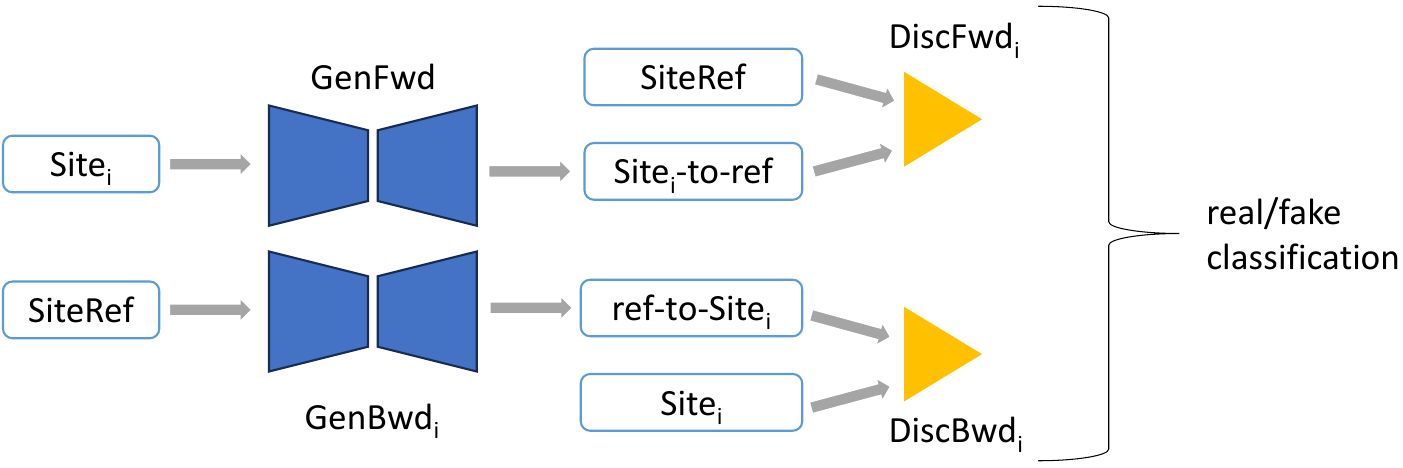}
        \caption{Training of the discriminators}
        \label{fig:iguane_training_disc}
    \end{subfigure}
    
    \begin{subfigure}{\linewidth}
        \centering
        \includegraphics[scale=0.4]{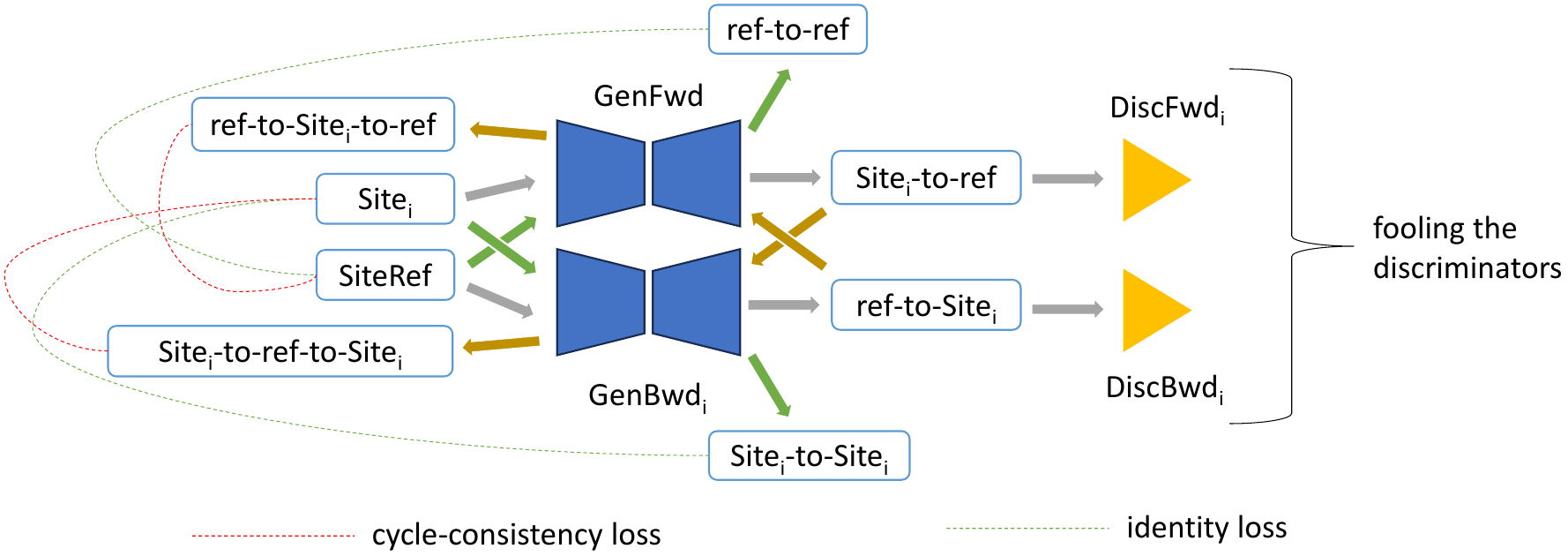}
        \caption{Training of the generators}
        \label{fig:iguane_training_gen}
    \end{subfigure}
    \caption{\textbf{Illustration of a training step in IGUANe.} At each step, the procedure is executed for each source site $Site_i$ ($i \in \{1,2,\ldots,N\}$, iteration in random order). In \ref{fig:iguane_training_disc}, (i) 2 $SiteRef$ images and 2 $Site_i$ images (translated to $SiteRef$ with $GenFwd$) are used to update $DiscFwd_i$, and (ii) 2 other $Site_i$ images and 2 other $SiteRef$ images (translated to $Site_i$ with $GenBwd_i$) are used to update $DiscBwd_i$. In \ref{fig:iguane_training_gen}, $GenFwd$ and $GenBwd_i$ are updated with an image from $SiteRef$ and another from $Site_i$ using adversarial, cycle-consistency and identity losses.}
    \label{fig:iguane_training}
\end{figure}

The training procedure is illustrated in Fig. \ref{fig:iguane_training}. Mean squared error is used as the adversarial loss \citep{Mao2017}. To generate realistic $SiteRef$ MR images, $GenFwd$ is successively trained adversarially with the $N$ forward discriminators. The $N$ backward generators are trained in the same way with each backward discriminator. To preserve original image information and regulate the generator trainings, cycle-consistency and identity losses are used \citep{Zhu2017}. The cycle-consistency losses are calculated by having $GenFwd$ collaborate with each backward generator through translation and inverse translation in both directions.

\subsubsection{Network architectures}
\label{sec:network_architectures}
For the generators, we followed the architecture described by \citet{Dewey2019}, which incorporates multiple skip connections to favor the conservation of anatomical details. A notable modification we have made is the use of 3D convolutions so that the model processes whole brain images. Additionally, for enhanced preservation of the original images’ content, we changed the task of the generator networks towards residual learning, which means that the generator output is added to the input image to obtain the final harmonized image \citep{deBel2021}. For the discriminators, we used patchGAN architectures \citep{Isola2017} with 3D convolutions and a receptive field of $54^3$ voxels.

We made some adjustments to better account for the fact that background represents a substantial portion of each image. First, the background intensity, which by default corresponds to the minimal intensity value in the image, is modified before feeding each image to the generators and discriminators. Specifically, it is set to the median brain intensity, which is the same for all images after preprocessing (section \ref{sec:preproc}). Our aim is to assign a more neutral value to a portion of the image that, despite its large size, should not influence the harmonization process. \citet{Robinson2020} implemented a similar idea. Additionally, the original brain mask is applied after each image translation to guide the generators to focus on brain intensities during training \citep{Roca2023}.

\subsubsection{Implementation}
\label{sec:iguane_implementation}
We trained IGUANe using the Training dataset, choosing SALD as the reference site due to its large number of MR images and a wide age range (19-80 years). The remaining ten sites in the Training dataset served as the source sites. Given the age imbalance between SALD and each source site (Table \ref{tab:training_dataset}), we implemented a sampling strategy for each source site to achieve a balanced age distribution \citep{Roca2023} with SALD. A probability distribution, based on participant ages, was thus effective for sampling MR images during the training sub-steps of each source site. The probability distributions used in this study can be visualized in \ref{appendix_probaSampling}.

To further promote the preservation of biological information, we set up a validation procedure that evaluated the current model every 5 epochs and saved the best version. For this purpose, two deep learning models pretrained for age and sex prediction using MR images from $SiteRef$ were applied to a subset of MR images from each source site harmonized towards SiteRef. The network architecture described by \citet{Cole2017} was used.

In this study, training IGUANe involved optimizing 11 generators and 20 discriminators simultaneously, imposing an important computational resource cost. To save GPU memory and speed-up the computations, a mixed precision policy was adopted \citep{Micikevicius2018}, allowing training on an NVIDIA Quadro RTX 6000 GPU with 24 GB of memory.

\subsection{Comparison with reference methods}
\subsubsection{Normalization techniques}
We implemented two widely adopted approaches in MR multicenter studies: histogram matching (HM) \citep{Bashyam2022,Fortin2016,Palladino2020,Robitaille2012,Wrobel2020} with the SALD dataset for establishing the standard scale to which the intensity histograms were matched, and WhiteStripe normalization (WS) \citep{Fortin2016,Gao2019,Wrobel2020}. We used the algorithms of \citet{Shah2011} and \citet{Shinohara2014} for HM and WS, respectively. For both methods, we preprocessed the MR images as done for IGUANe (section \ref{sec:preproc}) and used the brain masks to compute the statistics of intensity distributions.

\subsubsection{Style transfer approaches}
\label{sec:meth_styleTransfer}
We compared the performance of IGUANe to two style transfer approaches using the code and pretrained model available online: STGAN \citep{Liu2023}\footnotemark[10] and CALAMITI \citep{Zuo2021,Zuo2021_2}\footnotemark[11].

For STGAN, we used the reference image from the online repository for the style code extraction and as the harmonization target. Additionally, we performed intensity rescaling, not explicitly mentioned in the paper but found effective in the online code, as part of the preprocessing steps described in the study.

For CALAMITI, the reference image from the online repository served as the harmonization target. We followed the preprocessing steps given online. We used FSL-FLIRT with six degrees of freedom and a 0.8 mm$^3$ MNI template for registration. After the WS normalization, the last preprocessing step of CALAMITI, we scaled/shifted the MR intensities to align with the mean and the standard deviation of the normal appearing white-matter of the reference image (further details in \ref{appendix_calimiti_norm}).

\subsection{Experiments}
\label{sec:experiments}

\subsubsection{Harmonization on traveling subjects}
\label{sec:meth_travSubs}
We used the images from the Traveling subject dataset to assess the ability of harmonization models to transform images from one site to their equivalent in another.

For each subject, we computed a structural similarity index (SSIM) \citep{Wang2004} for each image pair, corresponding to each site pair. The intensity range for SSIM was set as the 99th percentile of voxel intensities from the two MR images. Since SSIM is designed for non-negative images \citep{Gourdeau2022,Wang2004}, we added constant intensities before performing computations on certain datasets (\ref{appendix_ssim_shifts}).

For each harmonization method, we assessed the statistical significance of the differences in SSIM before and after harmonization using a two-tailed Wilcoxon signed-rank test for clustered data \citep{Rosner2006}, treating each subject as a cluster. We then corrected the 5 p-values using the Benjamini-Hochberg procedure.

Additionally, we assessed the preservation of inter-subject differences after the harmonization process. Within each site, we computed the Euclidean distance for each image pair (corresponding to each subject pair) before and after harmonization. Next, we computed, for each site, the Pearson correlation coefficient between the distances before and after harmonization.

\subsubsection{Correlation between age and gray-matter volume}
\label{sec:meth_corrGMage}
In healthy adults, brain aging is characterized by a linear decrease in gray matter (GM) volume \citep{Ge2002,Hedman2012,Terribilli2011,Watanabe2013}. To study this pattern, we segmented the MR images from the Generalization dataset using SPM12 software and calculated the GM volume for each image, divided by the total intracranial volume. We then used the Pearson correlation coefficient to assess the linearity of GM loss with age. Additionally, we determined the slope of the linear least-squares regression of GM volume against age to quantify the rate of GM atrophy. A steeper slope indicates a more pronounced decrease in GM volume with age.

For each harmonization method, we assessed the statistical significance of the differences in correlations before and after harmonization using a two-tailed Steiger’s test \citep{Steiger1980} applied to the weighted correlation coefficients \citep{Costa2011}, where each participant’s weights sum to 1. We used the number of participants as sample size for the Steiger’s test. We then corrected the 5 p-values using the Benjamini-Hochberg procedure.

\subsubsection{Hippocampal volumes: case/control effect size}
\label{sec:meth_hippoVols}
The accelerated loss of hippocampal volume is a well-known pattern in Alzheimer’s disease \citep{Schuff2009}. To investigate this pattern, we randomly selected 250 CN and 250 AD participants from the Clinical dataset (mean age 71.34 and 71.90 in the CN and AD groups, respectively). Then, we segmented the MR images using SynthSeg \citep{Billot2023} and computed Cohen’s d scores between the hippocampal volumes in the CN and the AD groups.

\subsubsection{Prediction tasks}
\label{sec:meth_prediction}
We assessed the impact of harmonization on two prediction tasks: brain age and CN/AD classification. Brain age prediction involves training a model to predict an individual’s age from brain MRI data and has been widely investigated with deep learning models \citep{Cole2017,Gautherot2021,Jonsson2019}. Similarly, these approaches have been applied to classification of AD \citep{Basaia2019,Vieira2017}.

We implemented brain age prediction and CN/AD classification based on MR images, using the network architecture proposed by \citet{Cole2017}. For training the age prediction models, we randomly sampled 2178 MR images from the Training dataset, maintaining the proportion from each site. Evaluations were performed on MR images from the Generalization dataset. The CN/AD classifiers were trained using the AD\_train dataset and evaluated on the AD\_TEST, AD\_GE and MIRIAD datasets. Further details are given in \ref{appendix_predictive_models}.

We excluded STGAN and CALAMITI from these experiments due to the larger size of the MR images (almost $\times 3.4$), which would have implied much larger models, making training challenging. Thus, we trained and evaluated a brain age model and a CN/AD classifier on the preprocessed MR images (IGUANe preprocessing) as well as on the MR images after IGUANe, HM and WS harmonization. This resulted in the training and evaluation of 4 brain age models and 4 CN/AD classifiers.

For each harmonization method, we assessed the statistical significance of the differences in age prediction errors before and after harmonization using a two-tailed Wilcoxon signed-rank test for clustered data \citep{Rosner2006}, treating each subject as a cluster. We then corrected the 3 p-values with the Benjamini-Hochberg procedure.

\subsubsection{Ablation studies}
\label{sec:meth_ablationStudies}
We set up two variants of the IGUANe model in order to assess the impact of specific components on the preservation of biological information. In the first variant, we replaced the $N$ forward discriminators (section \ref{sec:universal_generator}) with a single discriminator trained to distinguish between real \textit{SiteRef} images and images from any of the $N$ source sites harmonized towards \textit{SiteRef} using \textit{GenFwd}. This allowed us to evaluate the influence of modifying the many-to-one training strategy proposed by \citet{Gao2019}. In the second variant, we kept this single forward discriminator but we additionally replaced the age-based sampling strategy (section \ref{sec:iguane_implementation}) by a classical equiprobable sampling scheme.

To compare these two variants with IGUANe, we studied the preservation of age information by computing the correlation between age and GM volume (section \ref{sec:meth_corrGMage}) and by evaluating the performance of brain age prediction (section \ref{sec:meth_prediction}).

\section{Results}
\subsection{Visual assessment of harmonization}
\label{sec:visua_harmonization}
\begin{figure}
    \centering
    \begin{subfigure}{\linewidth}
        \centering
        \includegraphics[scale=1]{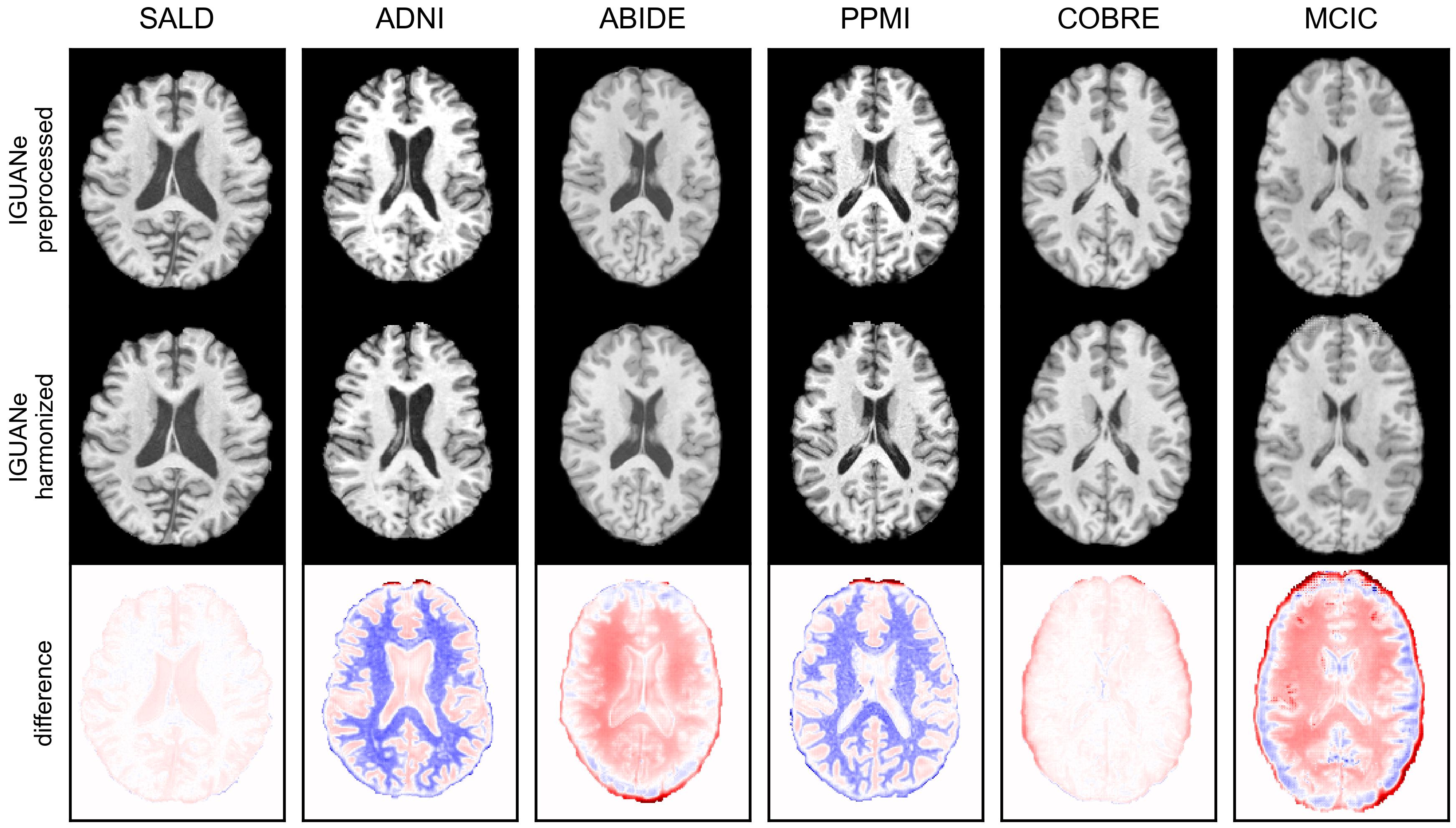}
    \end{subfigure}
    \begin{subfigure}{0.48\linewidth}
        \centering
        \includegraphics[scale=1]{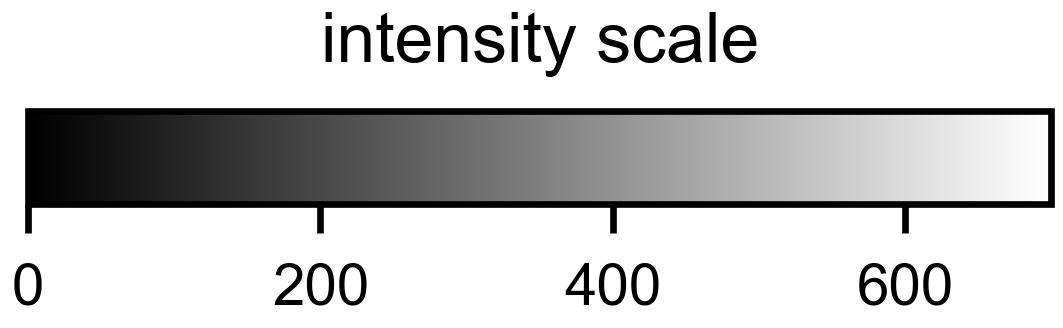}
    \end{subfigure}
    \begin{subfigure}{0.48\linewidth}
        \centering
        \includegraphics[scale=1]{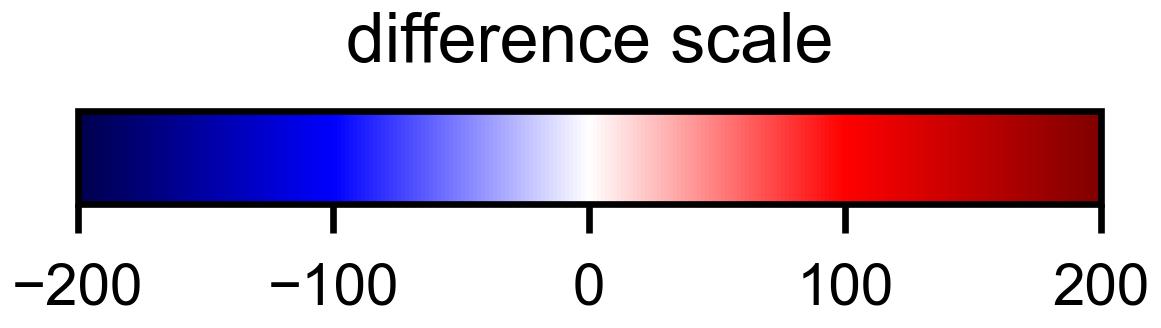}
    \end{subfigure}
    
    \caption{\textbf{Visualization of IGUANe harmonizations.} One image from the reference domain (SALD) and one from each study in the Generalization dataset were randomly sampled, and the middle axial slice is shown. The difference maps correspond to voxel-wise subtractions, i.e., the harmonized images minus the original.}
    \label{fig:slices_visu}
\end{figure}

The effect of IGUANe harmonization is illustrated in Fig. \ref{fig:slices_visu}. Since we designed the model for application to T1w images from any acquisition source, with a primary goal of preserving anatomical information, contrast modifications are not very visible. They were more obvious with the other implemented methods (\ref{appendix_slices_allMeths}). Nonetheless, the difference maps reveal that IGUANe can apply various modifications to input contrasts based on the input image. For example, IGUANe reduced the GM/white-matter contrast in the ADNI and PPMI images, while it increased it in the ABIDE and MCIC images.

It is also worth noting that the difference was slight for the SALD image, which was expected since SALD was the reference domain. Additional results supporting this are presented in \ref{appendix_results_sald}.

\subsection{Traveling subjects evaluations}
\label{sec:res_trav_subs}
Table \ref{tab:ssim} presents the SSIM values obtained from MR images of the same subject in the Traveling subject dataset. IGUANe harmonization showed negligible change, while SSIM increased slightly but significantly with HM, WS, and STGAN. A significant decrease was observed with CALAMITI. The differences in the SSIM values obtained after the preprocessing and before harmonization should also be noted. 

\begin{table}
    \setlength{\tabcolsep}{1pt}
    \centering
    \small
    \caption{\textbf{Structural similarity index (SSIM) in the Traveling subject dataset.}}
    \begin{tabular}{>{\raggedright}p{0.06\linewidth}>{\raggedright}p{0.12\linewidth}|>{\raggedright}p{0.16\linewidth}>{\raggedright}p{0.16\linewidth}>{\raggedright}p{0.16\linewidth}>{\raggedright}p{0.16\linewidth}>{\raggedright\arraybackslash}p{0.16\linewidth}}
         \hline
         &&HM&WS&IGUANe&STGAN&CALAMITI\\
         \cline{3-7}
         \multirow[c]{2}{*}{SSIM$^\#$}&preprocessed&0.872 ± 0.058&0.872 ± 0.058&0.872 ± 0.058&0.927 ± 0.024&0.761 ± 0.062\\
         &harmonized$^\dag$&+0.016 ± 0.015$^{**}$&+0.006 ± 0.033$^*$&+0.000 ± 0.006&+0.011 ± 0.008$^{**}$&-0.081 ± 0.064$^*$\\
         \hline
         \multicolumn{7}{p{\linewidth}}{$^\#$ Values are expressed as mean $\pm$ standard deviation.}\\
         \multicolumn{7}{p{\linewidth}}{$^\dag$ SSIM evolutions are indicated. Asterisks indicate significant Wilcoxon signed-rank tests comparing the SSIMs before and after harmonization (*: p$<$0.05; **: p$<$0.01; ***: p$<$0.001).}\\
    \end{tabular}
    \label{tab:ssim}
\end{table}

\begin{figure}
    \centering
    \begin{subfigure}{\linewidth}
        \centering
        \includegraphics[scale=1]{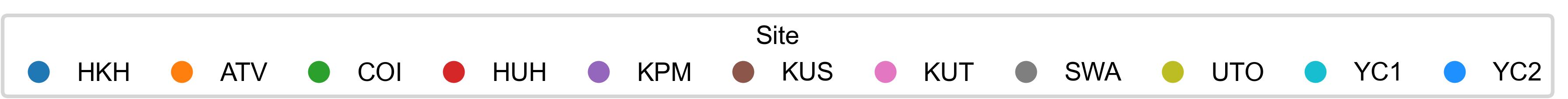}
    \end{subfigure}
    \par\bigskip

    \begin{subfigure}{0.19\linewidth}
        \centering
        \includegraphics[scale=1]{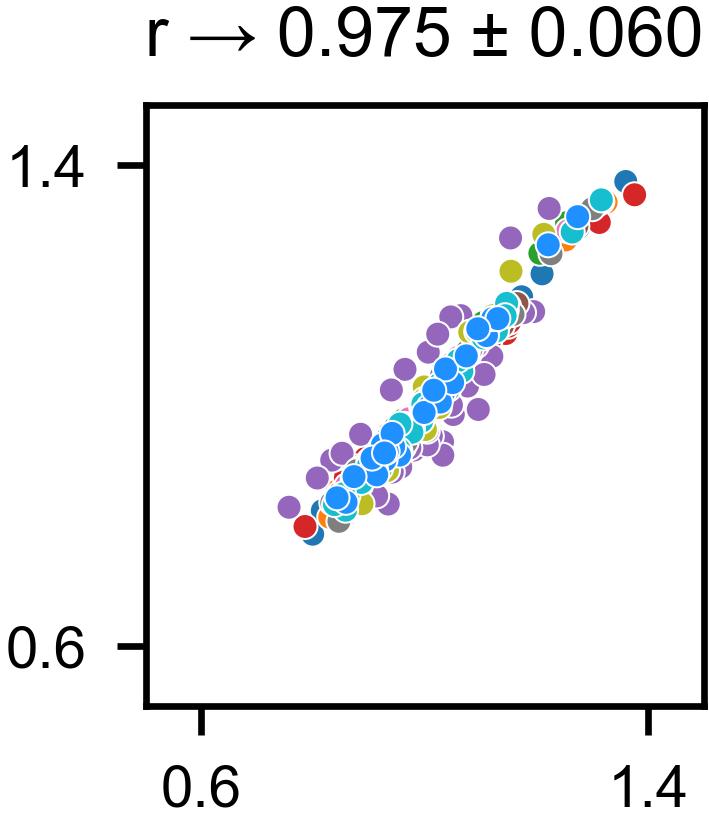}
        \caption{HM}
    \end{subfigure}
    \hfill
    \begin{subfigure}{0.19\linewidth}
        \centering
        \includegraphics[scale=1]{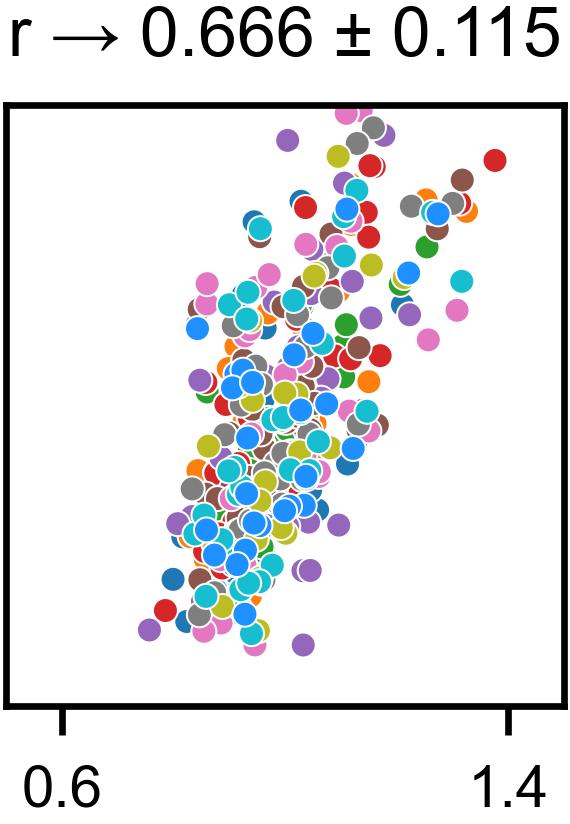}
        \caption{WS}
    \end{subfigure}
    \hfill
    \begin{subfigure}{0.19\linewidth}
        \centering
        \includegraphics[scale=1]{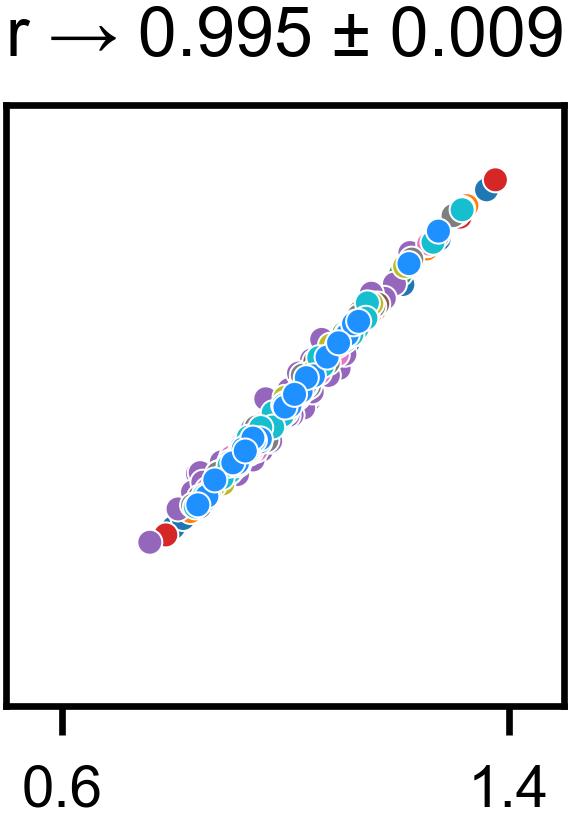}
        \caption{IGUANe}
    \end{subfigure}
    \hfill
    \begin{subfigure}{0.19\linewidth}
        \centering
        \includegraphics[scale=1]{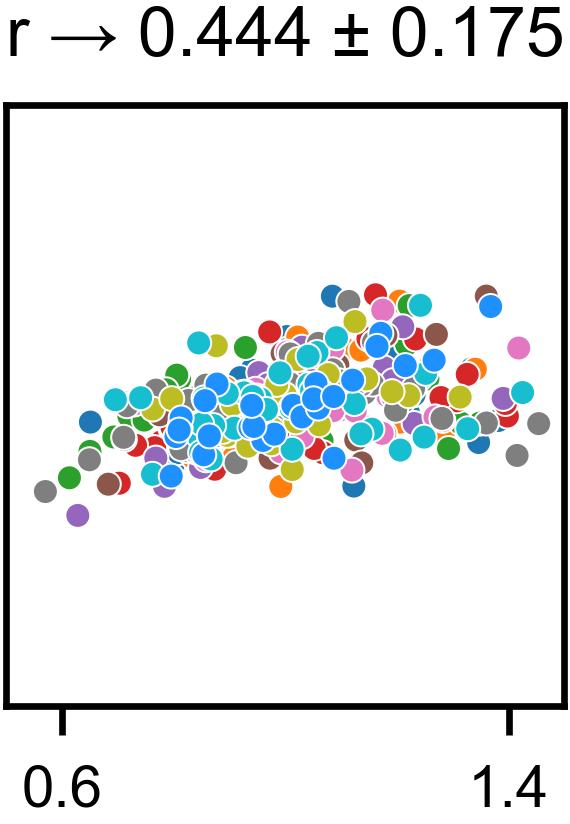}
        \caption{STGAN}
    \end{subfigure}
    \hfill
    \begin{subfigure}{0.19\linewidth}
        \centering
        \includegraphics[scale=1]{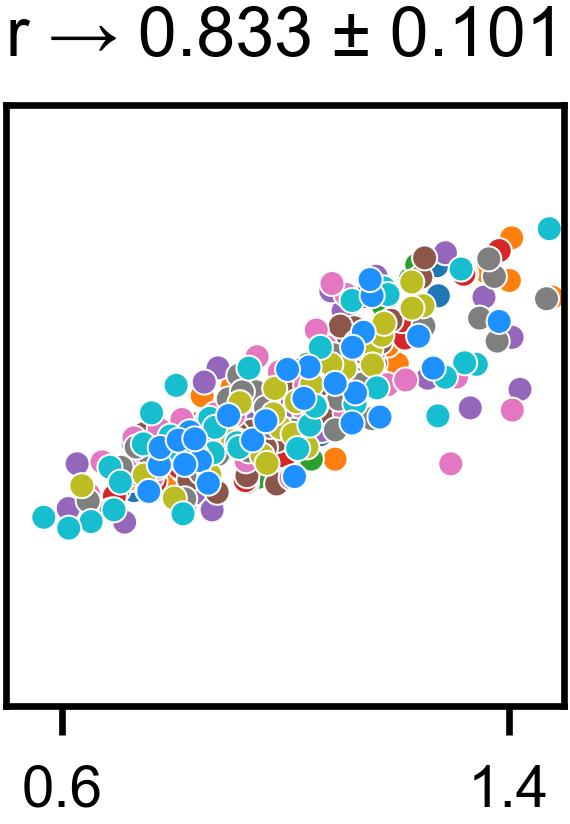}
        \caption{CALAMITI}
    \end{subfigure}
    
    \caption{\textbf{Inter-subject distances in each site of the Traveling subject dataset before and after harmonization.} The X and Y axes indicate the Euclidean distances divided by the average distance within each site before and after harmonization, respectively. The Pearson correlation coefficients (r) are expressed as mean $\pm$ standard deviation.}
    \label{fig:interSub_diffs}
\end{figure}

On the other hand, the intra-site correlation between the image distances before and after harmonization was much greater with IGUANe compared to the other methods (Fig. \ref{fig:interSub_diffs}).

\subsection{Correlation between age and gray-matter volume}
\label{sec:res_corr_gmAge}
\begin{figure}
    \centering
    \begin{subfigure}{\linewidth}
        \centering
        \includegraphics[scale=1]{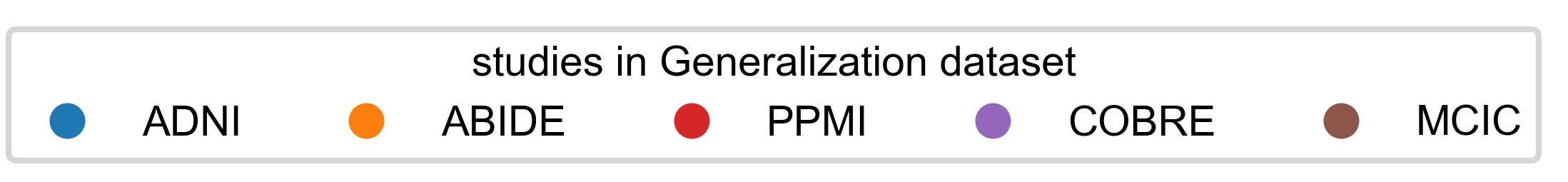}
    \end{subfigure}
    \par\bigskip

    \begin{subfigure}{0.22\linewidth}
        \centering
        \begin{minipage}{0.13\linewidth}
            \rotatebox[origin=center]{90}{before harmonization}
        \end{minipage}
        \begin{minipage}{0.82\linewidth}
            \includegraphics[scale=1]{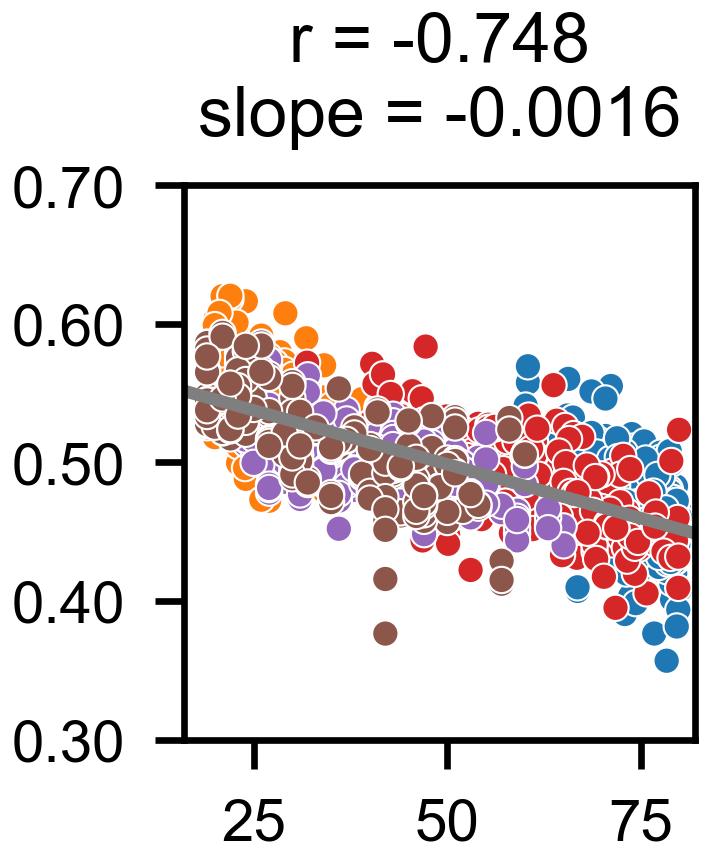}
        \end{minipage}

        \begin{minipage}{0.13\linewidth}
            \rotatebox[origin=center]{90}{after harmonization}
        \end{minipage}
        \begin{minipage}{0.82\linewidth}
            \includegraphics[scale=1]{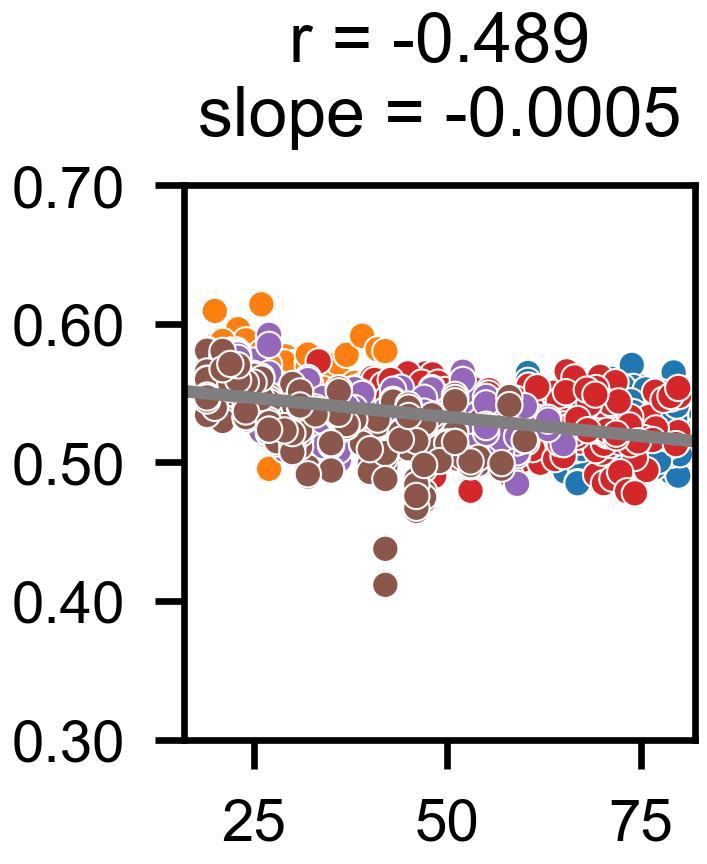}
        \end{minipage}
        \caption{HM}
        \label{fig:corr_gmAge_hm}
    \end{subfigure}
    \hfill
    \begin{subfigure}{0.18\linewidth}
        \centering
        \includegraphics[scale=1]{fig5abc_preproc.jpg}
        \includegraphics[scale=1]{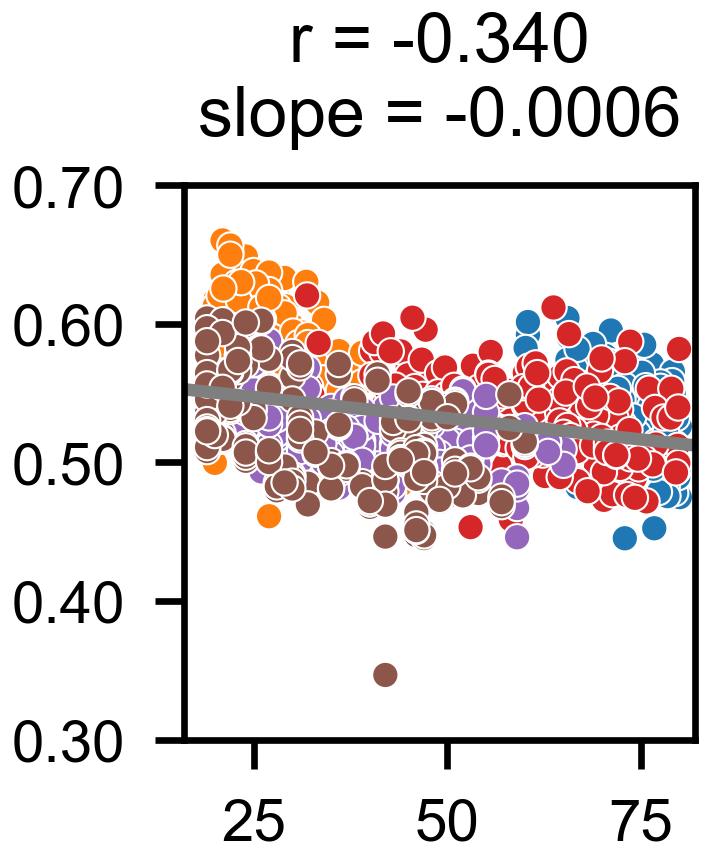}
        \caption{WS}
        \label{fig:corr_gmAge_ws}
    \end{subfigure}
    \hfill
    \begin{subfigure}{0.18\linewidth}
        \centering
        \includegraphics[scale=1]{fig5abc_preproc.jpg}
        \includegraphics[scale=1]{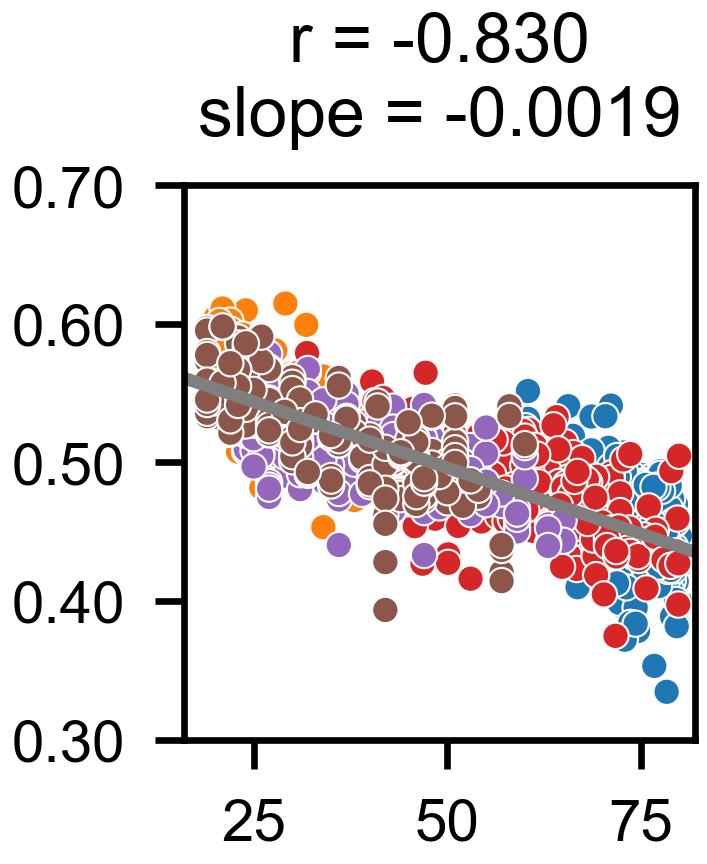}
        \caption{IGUANe}
        \label{fig:corr_gmAge_iguane}
    \end{subfigure}
    \hfill
    \begin{subfigure}{0.18\linewidth}
        \centering
        \includegraphics[scale=1]{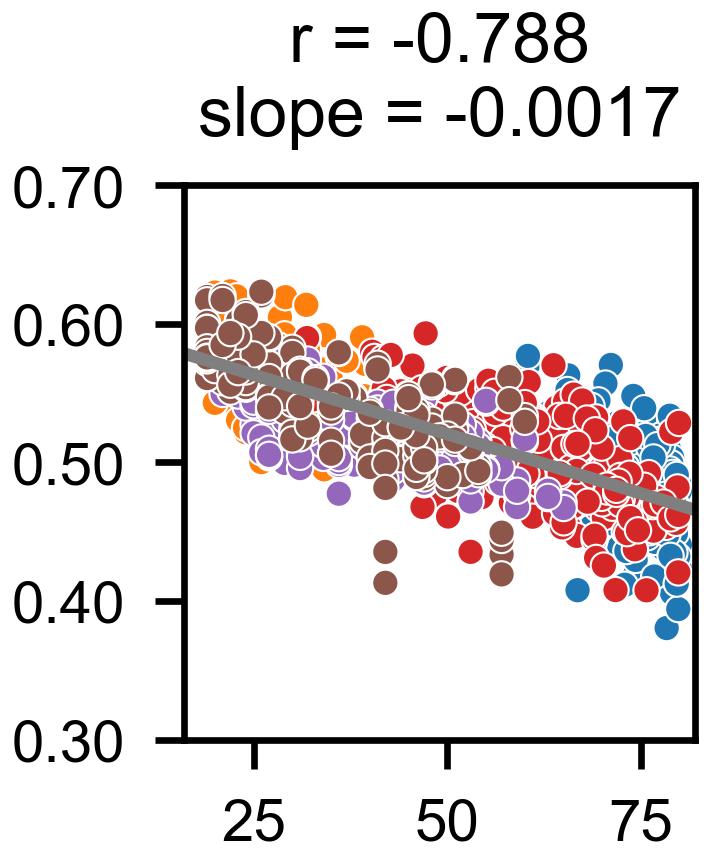}
        \includegraphics[scale=1]{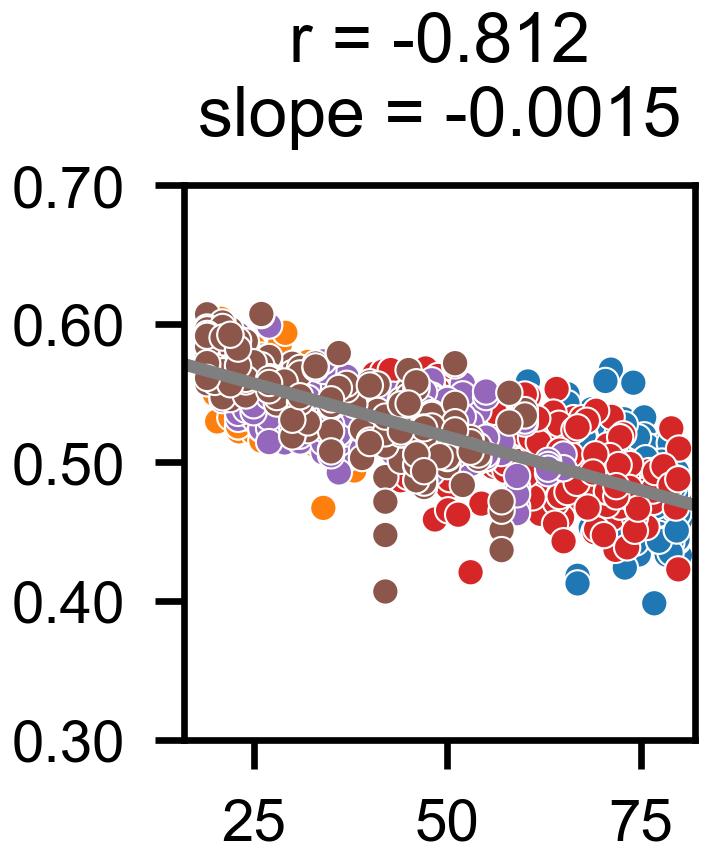}
        \caption{STGAN}
        \label{fig:corr_gmAge_stgan}
    \end{subfigure}
    \hfill
    \begin{subfigure}{0.18\linewidth}
        \centering
        \includegraphics[scale=1]{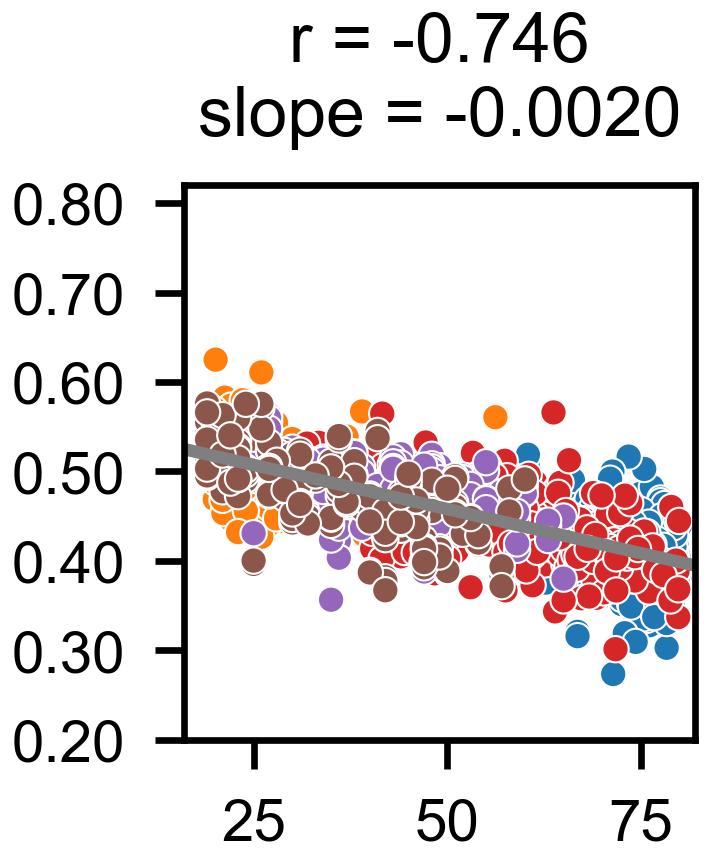}
        \includegraphics[scale=1]{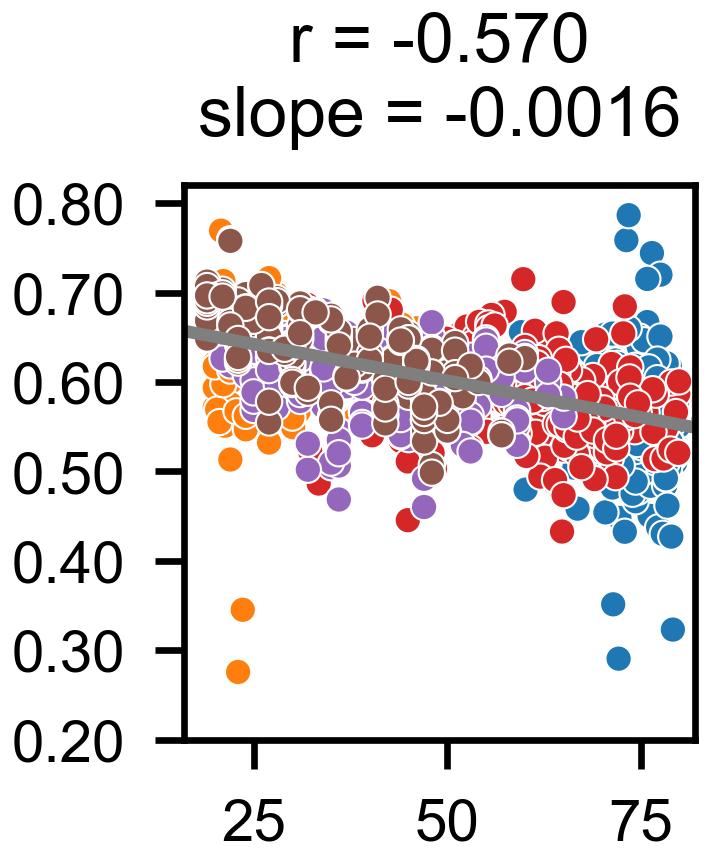}
        \caption{CALAMITI}
        \label{fig:corr_gmAge_calamiti}
    \end{subfigure}
    \caption{\textbf{Correlation between age and gray-matter (GM) volume in the Generalization dataset.} The X and Y axes correspond to ages and GM volumes (divided by the total intracranial volume), respectively. The linear least-squares regression line is plotted on each subfigure. The Steiger’s test comparing the correlation before and after harmonization yielded highly significant results (p$<$0.001) across all methods.}
    \label{fig:corr_gmAge}
\end{figure}

In Fig. \ref{fig:corr_gmAge}, it can be seen that the stronger negative correlation was achieved with IGUANe harmonization. Furthermore, IGUANe strengthened the regression slope (Fig. \ref{fig:corr_gmAge_iguane}), indicating a reinforcement of the GM loss pattern with harmonization. While STGAN also yielded a high correlation (Fig. \ref{fig:corr_gmAge_stgan}), it was weaker than IGUANe, and the regression slope was reduced. In contrast, HM, WS and CALAMITI diminished linearity and weakened the regression slope (Fig. \ref{fig:corr_gmAge_hm}, \ref{fig:corr_gmAge_ws} and \ref{fig:corr_gmAge_calamiti}, respectively). It should also be noted that the slope was steeper after CALAMITI preprocessing than after IGUANe harmonization; however, the correlation was clearly less linear according to the Pearson coefficient.

As most of MR image resolutions in the training dataset are around 1mm$^3$, we carried out supplementary analyzes of GM atrophy to study the value of IGUANe harmonization when applied to MR images with different resolutions. The results show that the pattern was better preserved with image resolutions around 1mm$^3$, but it was still reinforced at the multicenter level with harmonization (\ref{appendix_corrGMage_mcic}).

Additionally, a stronger negative correlation and a steeper regression slope were obtained with IGUANe compared to its ablated versions (\ref{appendix_ablation}).

\subsection{Comparison of hippocampal volumes}
\label{sec:res_hippoVols}
\begin{figure}
    \centering
    \includegraphics{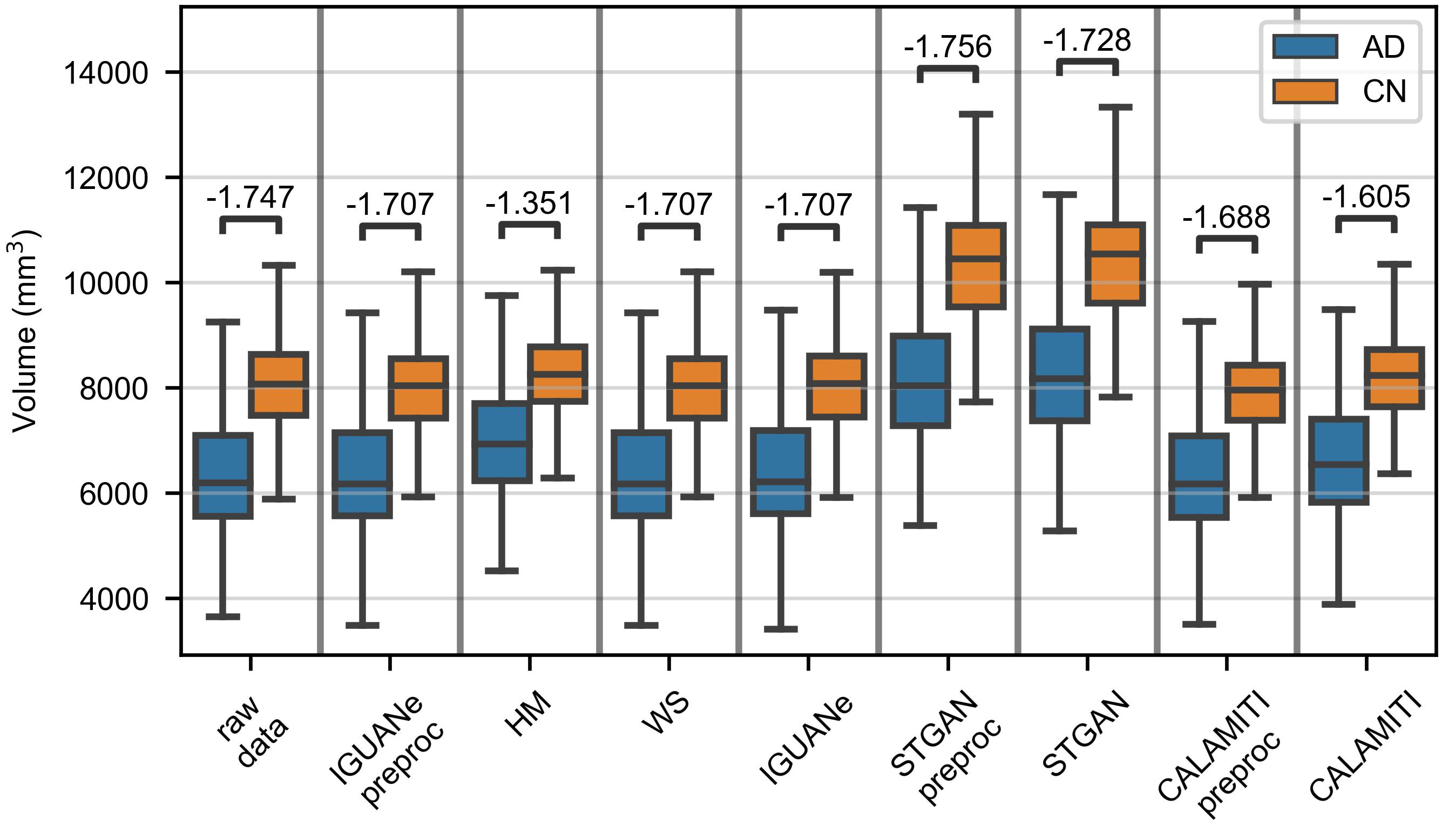}
    \caption{\textbf{Hippocampal volumes and comparison between healthy participants (CN) and subjects with Alzheimer’s disease (AD) in the Clinical dataset.} \textit{preproc} refers to the data obtained after preprocessing for the corresponding harmonization approach. Cohen’s d scores comparing the CN and AD groups are above the boxplots.}
    \label{fig:hippo_vols}
\end{figure}

Several observations emerge from the comparisons between hippocampal volumes of AD and CN participants in the Clinical dataset (Fig. \ref{fig:hippo_vols}). Firstly, the measures were higher for MR images preprocessed with STGAN compared to all other measures. Secondly, only WS and IGUANe preserved the effect size between the CN group and the AD group. Lastly, excluding STGAN, which produced different results, the most substantial effect size was obtained from raw data.

Supplementary results related to the case/control distinction in a single-scanner configuration are presented in \ref{appendix_hippocamp}. Additionally, an investigation into the reason behind the volume over-estimation in the STGAN preprocessing is carried out.

\subsection{Brain age prediction}
\label{sec:res_brainAge}
\begin{figure}
    \centering
    \begin{subfigure}{0.48\linewidth}
        \centering
        \includegraphics[scale=1]{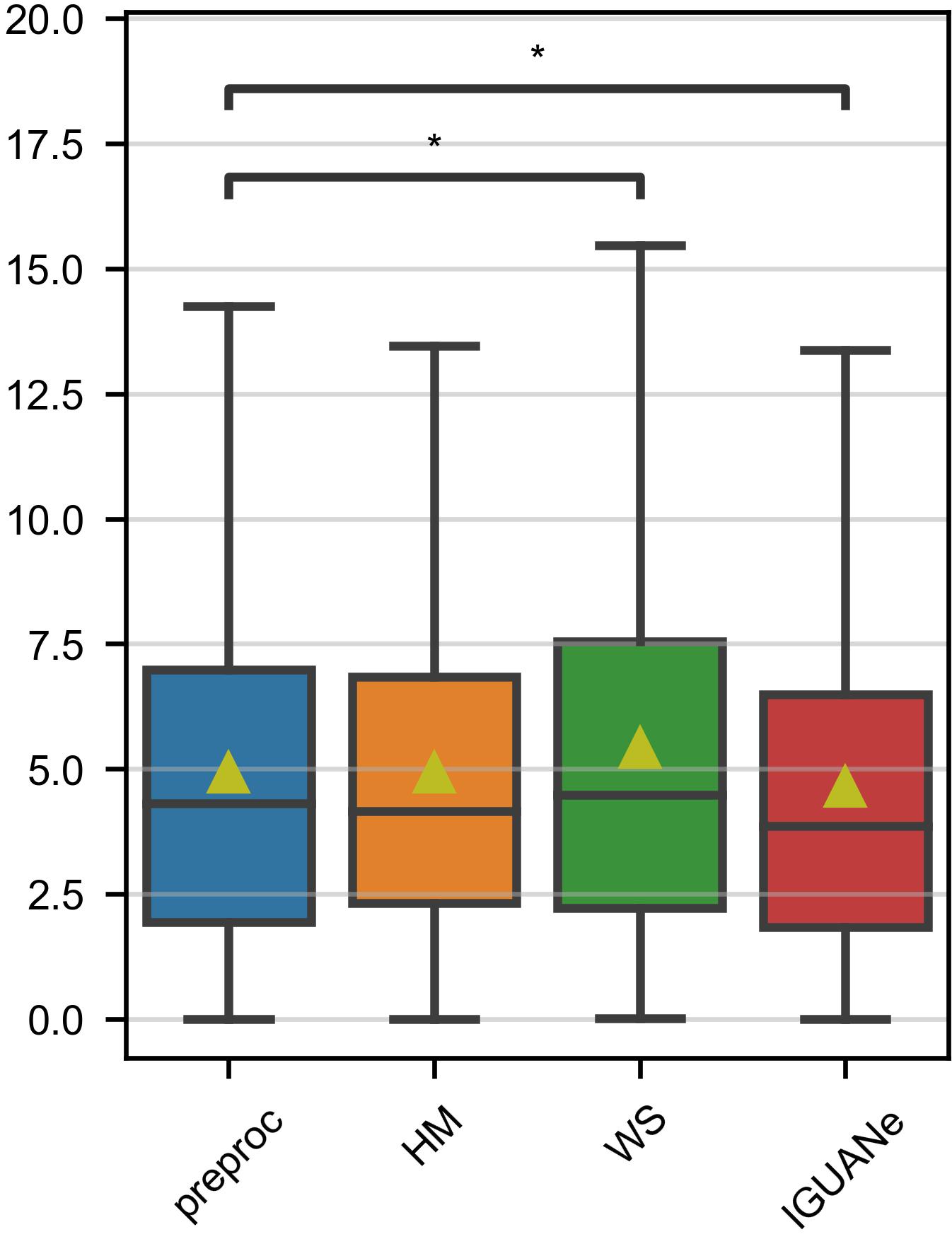}
        \caption{absolute errors}
        \label{fig:brainAge_errs}
    \end{subfigure}
    \hfill
    \begin{subfigure}{0.48\linewidth}
        \centering
        \includegraphics[scale=1]{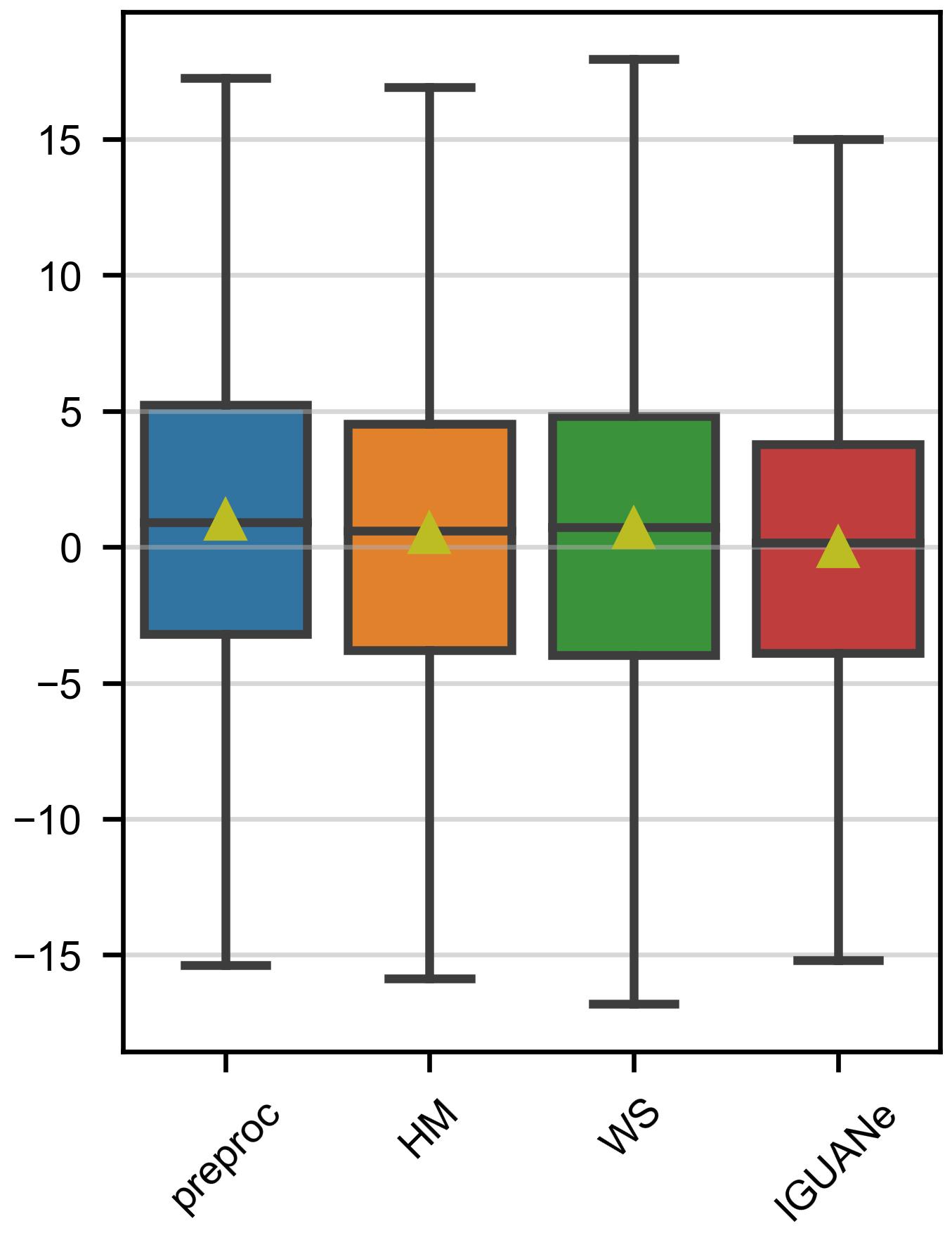}
        \caption{predicted age differences}
        \label{fig:brainAge_diffs}
    \end{subfigure}
    \caption{\textbf{Brain age prediction in the Generalization dataset.} \textit{preproc} refers to the images obtained after the IGUANe preprocessing. The triangles indicate the means. In \ref{fig:brainAge_errs}, asterisks indicate significant Wilcoxon signed-rank tests comparing the errors before and after harmonization (*: p$<$0.05; **: p$<$0.01; ***: p$<$0.001) . In \ref{fig:brainAge_diffs}, predicted age difference is computed as the predicted age minus the real age.}
\end{figure}

In Fig. \ref{fig:brainAge_errs}, it can be seen that brain age prediction was significantly improved with IGUANe (mean absolute error (MAE) decreased from 4.92 to 4.63). HM did not change the precision (MAE = 4.92), while WS led to a significant increase in errors (MAE = 5.42). A slight over-estimation pattern was observed in preprocessed data and after HM and WS, whereas the predicted age differences were more centered around zero after IGUANe (Fig. \ref{fig:brainAge_diffs}).

On the other hand, prediction performance was reduced with the ablated versions of IGUANe (\ref{appendix_ablation}).

\subsection{Classification of healthy and Alzheimer participants}
\label{sec:res_cn_ad}
The CN/AD classification results are presented in Table \ref{tab:cn_ad}. Across the three evaluation datasets, IGUANe achieved the highest accuracy and AUC score, except for the AUC score in the MIRIAD dataset. Overall, IGUANe demonstrated the best accuracy and AUC score, while only marginal improvements were observed with the other two methods.

\begin{table}
    \centering
    \caption{\textbf{Results in the classification of healthy and Alzheimer participants in the Clinical dataset.}}
    \begin{tabular}{l|ll|ll|ll|ll}
    \hline
         &\multicolumn{2}{l|}{preproc$^\#$}&\multicolumn{2}{l|}{HM}&\multicolumn{2}{l|}{WS}&\multicolumn{2}{l}{IGUANe}\\
         &accuracy&AUC&accuracy&AUC&accuracy&AUC&accuracy&AUC\\
         \hline
         AD\_TEST$^\dag$&0.843&0.901&0.809&0.926&0.834&0.924&\textbf{0.872}&\textbf{0.938}\\
         AD\_GE$^\dag$&0.828&0.894&0.837&0.920&0.835&0.902&\textbf{0.839}&\textbf{0.921}\\
         MIRIAD$^\dag$&0.880&0.977&0.920&0.984&0.931&\textbf{0.992}&\textbf{0.933}&0.977\\
         \hline
         GLOBAL$^\dag$&0.845&0.919&0.850&0.939&0.858&0.934&\textbf{0.871}&\textbf{0.942}\\
         \hline
         \multicolumn{9}{@{}l}{$^\#$\textit{preprocs} refers to the images obtained after the IGUANe preprocessing.}\\
         \multicolumn{9}{@{}l}{$^\dag$ Bold font indicates the best accuracies and AUC scores.}
    \end{tabular}
    \label{tab:cn_ad}
\end{table}

\section{Discussion}
We presented IGUANe, a model for inter-site harmonization of complete 3D MR images. It adopts a unified training procedure for exploiting images from multiple domains in parallel. Its original many-to-one learning framework allows for the harmonization of MR images from any site. Following a training phase with a dataset including 11 scanners for a total of 4347 T1w brain images, we applied the model to data from additional studies without any fine-tuning. Various experiments attested that, thanks to its learning approach based on domain pairs and a biased sampling strategy, IGUANe successfully preserved and enhanced patterns related to age and AD. Comparisons with two intensity normalization techniques and two style transfer methods demonstrated the robustness of the proposed approach.

\subsection{Signal quality}
The visualization of harmonized MR images (section \ref{sec:visua_harmonization}) and the SSIMs obtained with the traveling subjects (section \ref{sec:res_trav_subs}) indicate that IGUANe introduces minimal changes to the input images and may not fully homogenize contrasts. It could be inconvenient if subsequent analyses were very sensible to small contrast differences. However, analyzes conducted - notably the inter-subject differences (section \ref{sec:res_trav_subs}) - suggest that IGUANe better preserved relevant variability and that the SSIM increase with the other methods may be partly due to over-homogenization.

\citet{Liu2023} demonstrated effective preservation of inter-subject variabilities with STGAN, which contrasts with our results. This disparity may stem from their experiment being based on only 10 input MR images, all acquired with a Siemens 3 Tesla from the same study included in the training of STGAN. Our findings also highlight that voxel-level similarity metrics are significantly influenced by image preprocessing steps, such as resampling and skull-stripping. These factors should be taken into consideration when comparing methods with distinct preprocessing approaches. Our results align with previous studies that have underscored the limitations of SSIM in medical imaging \citep{Pambrun2015,Ravano2022}.

\subsection{Patterns of brain aging}
The analysis of the correlation between age and GM volume has the advantage of assessing potential reinforcements of a specific pattern of brain aging. IGUANe exhibited a strong linearity and a steep regression slope compared to other approaches (section \ref{sec:res_corr_gmAge}), suggesting that IGUANe could serve as a preliminary step in studies relying on automated segmentation software.

We further evaluated the preservation of aging patterns through brain age prediction. Brain age prediction is commonly employed to assess harmonization by training a model on MR images from one site and applying it to MR images from others \citep{Bashyam2022,Liu2023}. However, this setting is not common, and recent studies establishing brain age models often utilize large multicenter training sets, rendering models robust to site effects \citep{Bashyam2020,Cole2018,Gautherot2021}. In our case, employing such a training set (section \ref{sec:meth_prediction}), we demonstrated that even in this configuration, performances were enhanced after IGUANe harmonization, while they remained similar with HM and diminished with WS (section \ref{sec:res_brainAge}). It is important to note that these positive results should be tempered, as brain age prediction was optimized during the training phase of IGUANe harmonization using a subset of the training set.

The enhancement of aging patterns with IGUANe is a significant result, especially given the substantial age imbalances in the training set. As suggested by \citet{Zuo2021}, these imbalances may cause confusion between site-related variabilities and variabilities due to anatomical differences in harmonization models. To address this issue, we employed an age-based sampling strategy to equalize age distributions during the training phase. Its value has been demonstrated in a previous work \citep{Roca2023}, and in this study, we further validated its effectiveness by applying our model to data from unseen sites. The \citet{Cackowski2023}'s model uses a biological preservation module to mitigate overcorrection issues. However, their evaluation, which focuses on autism detection, does not consider performance under conditions of imbalance in the variable of interest within the training set, nor does it assess the model's performance when applied to images from external datasets.

\subsection{Patterns of Alzheimer's disease}
Despite exclusively including healthy participants in the training set of IGUANe to avoid confounding effects and overcorrection, the results indicate that IGUANe not only preserved age-related variabilities in healthy populations but also patterns associated with AD. Similar to the findings of \citet{Liu2023} with STGAN, IGUANe demonstrated the ability to retain differences in hippocampal volumes between CN and AD participants (section \ref{sec:res_hippoVols}). Moreover, IGUANe improved CN/AD classification, both with data originating from distributions similar to the classifier training set and with data acquired from unseen MRI scanners (GE manufacturer) and/or in different studies (section \ref{sec:res_cn_ad}).

While the CN/AD differentiation remained intact with IGUANe harmonization, the analysis of hippocampal volumes in relation to the case/control distinction revealed a more substantial effect size before any processing (section \ref{sec:res_hippoVols}). This suggests more consistent segmentations, which aligns with expectations given SynthSeg's design for MRI contrast independence \citep{Billot2023}. Consequently, this underscores that harmonization may not always be necessary depending on the subsequent analytical goals.

\subsection{Comparison with other harmonization methods}
Numerous methods have been proposed for harmonizing MR images from unseen sites, but the lack of standardized practices to ensure method reproducibility poses a challenge for meaningful comparisons. We opted to compare our approach with two others, STGAN and CALAMITI, for which both the code and the trained model were available. The issue of reproducibility has also been highlighted by \citet{Hu2023} in the context of harmonization models based on deep learning.

The results we obtained with CALAMITI may be influenced by potential challenges related to method reproducibility, as indicated by certain ambiguities identified in the preprocessing steps (section \ref{sec:meth_styleTransfer} and \ref{appendix_calimiti_norm}). It is noteworthy that while \citet{Zuo2021} used fine-tuning to adapt CALAMITI to new sites, their observation of favorable outcomes even without fine-tuning appears to differ from our findings.

\subsection{Limitations and perspectives}
In this study, our focus was on the harmonization of T1w images, a sequence widely used in research studies and clinical practice. The simplicity of IGUANe’s training and application, which exclusively requires T1w images, is noteworthy. Nevertheless, assessing the model’s ability to harmonize other MRI modalities could be valuable. Several options can be explored, including training a distinct model for each image type, amalgamating all images during the training phase of a single model, or incorporating input/output channels in the network architectures for the parallel harmonization of multiple sequences. \citet{Dewey2019} found that the latter approach improved their harmonization model.

To further assess the generalizability of our approach, additional experiments involving other pathologies could be conducted. For instance, exploring whether the harmonization of MR images with brain lesions would require retraining with more representative populations to prevent alteration of lesion characteristics is an avenue worth investigating.

\section{Conclusion}
This study introduces IGUANe, an unsupervised generative model designed for the inter-site harmonization of structural brain MR images. We present a model trained on a large multicenter dataset of T1w images, capable of seamlessly harmonizing images from any site. The IGUANe framework incorporates adversarial training among multiple learning modules and is specifically designed for harmonization without overcorrection. Our experimentations, conducted on diverse cohorts from multiple studies not seen during the training phase, demonstrate that IGUANe enhances aging patterns and differences between CN and AD participants. They also highlight the robustness of our approach compared to other harmonization methods, whether in terms of segmentation consistency or prediction performances. IGUANe holds promise for future multicenter studies, providing a tool for harmonizing images without the requirement of a new training phase.

\section*{Acknowledgements}
Data were provided in part by OASIS-3: Longitudinal Multimodal Neuroimaging: Principal Investigators: T. Benzinger, D. Marcus, J. Morris; NIH P30 AG066444, P50 AG00561, P30 NS09857781, P01 AG026276, P01 AG003991, R01 AG043434, UL1 TR000448, R01 EB009352. In the OASIS-3 study, AV-45 doses were provided by Avid Radiopharmaceuticals, a wholly owned subsidiary of Eli Lilly.

Data used in preparation of this article were obtained in part from the Neuromorphometry by Computer Algorithm Chicago (NMorphCH) dataset. As such, the investigators within NMorphCH contributed to the design and implementation of NMorphCH and/or provided data but did not participate in analysis or writing of this report.

Data collection and sharing for this project was funded in part by NIMH grant R01 MH056584.

Data were provided in part by the Human Connectome Project, WU-Minn Consortium (Principal Investigators: David Van Essen and Kamil Ugurbil; 1U54MH091657) funded by the 16 NIH Institutes and Centers that support the NIH Blueprint for Neuroscience Research; and by the McDonnell Center for Systems Neuroscience at Washington University.

Data collection and sharing for this project was provided in part by the International Consortium for Brain Mapping (ICBM; Principal Investigator: John Mazziotta, MD, PhD). ICBM funding was provided by the National Institute of Biomedical Imaging and BioEngineering. ICBM data are disseminated by the Laboratory of Neuro Imaging at the University of Southern California.

Data collection and sharing for this project was funded in part by ADNI (National Institutes of Health Grant U01 AG024904) and DOD ADNI (Department of Defense award number W81XWH-12-2-0012). ADNI is funded by the National Institute on Aging, the National Institute of Biomedical Imaging and Bioengineering, and through generous contributions from the following: AbbVie, Alzheimer’s Association; Alzheimer’s Drug Discovery Foundation; Araclon Biotech; BioClinica, Inc.; Biogen; Bristol-Myers Squibb Company; CereSpir, Inc.; Cogstate; Eisai Inc.; Elan Pharmaceuticals, Inc.; Eli Lilly and Company; EuroImmun; F. Hoffmann-La Roche Ltd and its affiliated company Genentech, Inc.; Fujirebio; GE Healthcare; IXICO Ltd.; Janssen Alzheimer Immunotherapy Research \& Development, LLC.; Johnson \& Johnson Pharmaceutical Research \& Development LLC.; Lumosity; Lundbeck; Merck \& Co., Inc.; Meso Scale Diagnostics, LLC.; NeuroRx Research; Neurotrack Technologies; Novartis Pharmaceuticals Corporation; Pfizer Inc.; Piramal Imaging; Servier; Takeda Pharmaceutical Company; and Transition Therapeutics. The Canadian Institutes of Health Research is providing funds to support ADNI clinical sites in Canada. Private sector contributions are facilitated by the Foundation for the National Institutes of Health (www.fnih.org). The grantee organization is the Northern California Institute for Research and Education, and the study is coordinated by the Alzheimer’s Therapeutic Research Institute at the University of Southern California. ADNI data are disseminated by the Laboratory for Neuro Imaging at the University of Southern California.

The imaging data and demographic information was collected and shared in part by the Mind Research Network supported by the Department of Energy under Award Number DE-FG02-08ER64581.

Data used in the preparation of this article were obtained in part on June 1, 2023 from the Parkinson’s Progression Markers Initiative (PPMI) database (\url{https://www.ppmi-info.org/access-data-specimens/download-data}), RRID:SCR\_006431. For up-to-date information on the study, visit \url{www.ppmi-info.org}.

PPMI – a public-private partnership – is funded by the Michael J. Fox Foundation for Parkinson’s Research and funding partners, including 4D Pharma, Abbvie, AcureX, Allergan, Amathus Therapeutics, Aligning Science Across Parkinson's, AskBio, Avid Radiopharmaceuticals, BIAL, BioArctic, Biogen, Biohaven, BioLegend, BlueRock Therapeutics, BristolMyers Squibb, Calico Labs, Capsida Biotherapeutics, Celgene, Cerevel Therapeutics, Coave Therapeutics, DaCapo Brainscience, Denali, Edmond J. Safra Foundation, Eli Lilly, Gain Therapeutics, GE HealthCare, Genentech, GSK, Golub Capital, Handl Therapeutics, Insitro, Janssen Neuroscience, Jazz Pharmaceuticals, Lundbeck, Merck, Meso Scale Discovery, Mission Therapeutics, Neurocrine Biosciences, Neuropore, Pfizer, Piramal, Prevail Therapeutics, Roche, Sanofi, Servier, Sun Pharma Advanced Research Company, Takeda, Teva, UCB, Vanqua Bio, Verily, Voyager Therapeutics, the Weston Family Foundation and Yumanity Therapeutics.

Data was downloaded in part from the COllaborative Informatics and Neuroimaging Suite Data Exchange tool (COINS; \url{http://coins.mrn.org/dx}) and data collection was performed at the Mind Research Network, and funded by a Center of Biomedical Research Excellence (COBRE) grant 5P20RR021938/P20GM103472 from the NIH to Dr. Vince Calhoun.

Data used in the preparation of this article were obtained in part from the MIRIAD database. The MIRIAD investigators did not participate in analysis or writing of this report. The MIRIAD dataset is made available through the support of the UK Alzheimer’s Society (Grant RF116). The original data collection was funded through an unrestricted educational grant from GlaxoSmithKline (Grant 6GKC).

\appendix
%\counterwithin{figure}{section}

\section{Image resolutions}
\label{appendix_resolutions}
\setcounter{table}{0}

The resolutions of all the images used in this study are listed in Table \ref{tab:resolutions}.

\begin{table}
    \centering
    \caption{\textbf{Voxel resolutions in LAS orientation in each dataset.}}
    \begin{tabular}{p{0.2\linewidth}p{0.7\linewidth}}
        \hline
         \textbf{Study / Dataset name}&\textbf{Resolution, mm$^3$}$^\#$\\
         \hline
         \multicolumn{2}{l}{\textbf{Training dataset}}\\
         SALD&1.0 x 1.0 x 1.0 (494)\\
         IXI 1.5 Tesla&1.2 x 0.9 x 0.9 (301); 1.2 x 1.0 x 1.0 (4)\\
         IXI 3 Tesla&1.2 x 0.9 x 0.9 (176)\\
         OASIS TrioTim&1.0 x 1.0 x 1.0 (854); 1.2 x 1.1 x 1.1 (2); 1.0 x 1.0 x 3.0 (1)\\
         OASIS Biograph&1.2 x 1.1 x 1.1 (382); 1.0 x 1.0 x 1.0 (30)\\
         NKI-RS&1.0 x 1.0 x 1.0 (249)\\
         NMorphCH&1.0 x 1.0 x 1.0 (139); 1.6 x 1.0 x 1.0 (2)\\
         AIBL&1.2 x 1.0 x 1.0 (348); 1.0 x 1.0 x 1.0 (140); 11.0 x 1.0 x 1.0 (1)\\
         HCP&0.7 x 0.7 x 0.7 (402)\\
         ICBM Sonata&1.0 x 1.0 x 1.0 (677)\\
         ICBM ACS III&1.0 x 1.0 x 1.0 (145)\\
         \hline
         \multicolumn{2}{l}{\textbf{Generalization dataset}}\\
         ADNI&1.2 x 0.9 x 0.9 (104); 1.2 x 1.0 x 1.0 (90); 1.2 x 1.1 x 1.1 (64); 1.0 x 1.0 x 1.0 (42); 1.2 x 1.2 x 1.2 (24); 1.0 x 1.1 x 1.1 (3); 0.5 x 1.0 x 0.5 (1)\\
         MCIC&0.6 x 1.5 x 0.6 (174); 0.7 x 1.5 x 0.7 (70)\\
         PPMI&1.0 x 1.0 x 1.0 (215); 1.2 x 0.9 x 0.9 (22); 1.2 x 1.1 x 1.1 (4)\\
         COBRE&1.0 x 1.0 x 1.0 (227)\\
         ABIDE&1.0 x 1.0 x 1.0 (67); 1.3 x 1.0 x 1.0 (27); 1.2 x 0.5 x 0.5 (21); 1.0 x 1.2 x 1.0 (14); 1.1 x 1.1 x 1.1 (10)\\
         \hline
         \multicolumn{2}{l}{\textbf{Clinical dataset}}\\
         AD\_train&1.2 x 1.0 x 1.0 (658); 1.2 x 1.1 x 1.1 (365); 1.2 x 1.2 x 1.2 (325); 1.0 x 1.0 x 1.0 (247); 1.2 x 0.9 x 0.9 (156); 1.2 x 1.3 x 1.3 (22); 1.2 x 1.4 x 1.4 (1); 1.0 x 1.1 x 1.1 (1)\\
         AD\_test&1.2 x 1.0 x 1.0 (244); 1.2 x 1.2 x 1.2 (143); 1.2 x 1.1 x 1.1 (127); 1.0 x 1.0 x 1.0 (111); 1.2 x 0.9 x 0.9 (49); 1.2 x 1.3 x 1.3 (11); 1.3 x 0.9 x 0.9 (1); 0.5 x 1.0 x 0.5 (1)\\
         AD\_GE&1.2 x 0.9 x 0.9 (643); 1.2 x 1.0 x 1.0 (375); 1.2 x 1.1 x 1.1 (147); 1.0 x 1.0 x 1.0 (113); 1.0 x 1.1 x 1.1 (17)\\
         MIRIAD&0.9 x 1.5 x 0.9 (652)\\
         \hline
         \multicolumn{2}{@{}l}{$^\#$The number of MR images is indicated in brackets.}\\
    \end{tabular}
    \label{tab:resolutions}
\end{table}

\section{Supplementary details of the IGUANe implementation}
\label{appendix_iguane}

\subsection{MRI cropping}
We cropped the MR images so that their dimensions went from 182 $\times$ 218 $\times$ 182 to 160 $\times$ 192 $\times$ 160 (section \ref{sec:preproc}). To this end, we computed for each MR image the number of background slices (with only zeros) on each of the six volume sides and cropped it according to the results. Images without enough background slices were excluded (e.g. 1.8\% in the Training dataset). Nonetheless, it should be noted that the trained model that we used and that we made available online can be applied to any image with dimensions divisible by 16 (e.g. 192 $\times$ 224 $\times$ 192). It can therefore be applied easily to any MR image with zero padding.

\subsection{Scaling/shifting of MRI intensities}
After preprocessing, the background and the median brain intensity are at 0 and 1, respectively (section \ref{sec:preproc}). The intensities are then multiplied by 500 to set the median brain intensity to 500.\\
Before the images are fed to IGUANe, the intensities are divided by 500 and decreased by 1 so that the background and the median brain intensity are at -1 and 0, respectively. After inference, the intensities are incremented by 1 and then multiplied by 500 to restore a similar intensity scale.\\
For the evaluations relying on deep learning models (section \ref{sec:meth_prediction}), the intensities are also divided by 500 and decreased by 1 (not for HM- and WS-normalized images, see \ref{appendix_predictive_models}).

\subsection{Generator loss}
A voxel-wise mean absolute difference is used for the cycle loss. Given an image $x$ and the cycled image $x^\prime$, the formula is the following: $L_{cyc} = \mid x-x^\prime \mid_1 / N$ with $N$ the number of voxels.\\
The identity loss is computed in the same way. Given an image $x$ and the image $x^\prime$ translated into the same domain, we have $L_{id} = \mid x-x^\prime \mid_1 / N$.\\
The global loss for each generator is then computed in the following way: $L = L_{adv} + \lambda \times L_{cyc} + \lambda / 2 \times L_{id}$ with $L_{adv}$ the adversarial loss. We set $\lambda$ to 30.

\subsection{Network architectures}
Let $x$, $y$, $z$ integers. CxSyKz a Convolution-InstanceNormalization-LeakyReLU block with $x$ filters, $z^3$ kernels and strides $y^3$. The discriminators are composed of 4 consecutive blocks: $$C64S2K4-C128S2K4-C256S2K4-C1S1K3$$.

The main components of the generator architectures are described in section \ref{sec:network_architectures}.  The last activation function before the addition is tanh, to enable both negative and positive residuals. During inference, negative voxels in the output volume are clipped to the background value.

Both the generators and discriminators use instance normalization \citep{Ulyanov2017} with mean absolute deviation instead of standard deviation \citep{Wu2019}.

\subsection{Training details}
In this work, we trained IGUANe for 100 epochs of 200 steps (the procedure of a training step is given in section \ref{sec:training_procedure}). We used an Adam optimizer \citep{Kingma2017} with a linear learning rate decay from 0.0002 to 0.00002. We set the batch size to 1 for the generators (one image for each of the two domains at each sub-step) and to 2 for the discriminators (two images for each of the two domains at each sub-step).

We implemented a data-augmentation that consisted in a random translation ($\pm$ 5 voxels) along the three orthogonal axes and a random rotation ($\pm$ 10 °) along a randomly selected orthogonal plane (rotation applied with a probability 1/2).

We randomly selected 44 participants from each source site of the Training dataset for the validation procedure (44 corresponds to the set with the least number of participants, i.e. NMorpchCH). Using the accuracy of sex prediction and the coefficient of determination (R2) of age prediction, we defined the following metric to determine and keep the best model: 0.75 $\times$ R2 + 0.25 $\times$ accuracy. Details about the predictive models used in this study are given in \ref{appendix_predictive_models}.

\section{Sampling probabilities for IGUANe training}
\label{appendix_probaSampling}
\setcounter{figure}{0}
Fig. \ref{fig:probas_sampling} illustrates the sampling strategy used for each source site of the Training dataset for IGUANe training.

\begin{figure}
    \centering
    \begin{subfigure}{0.48\linewidth}
        \centering
        \includegraphics[scale=0.9]{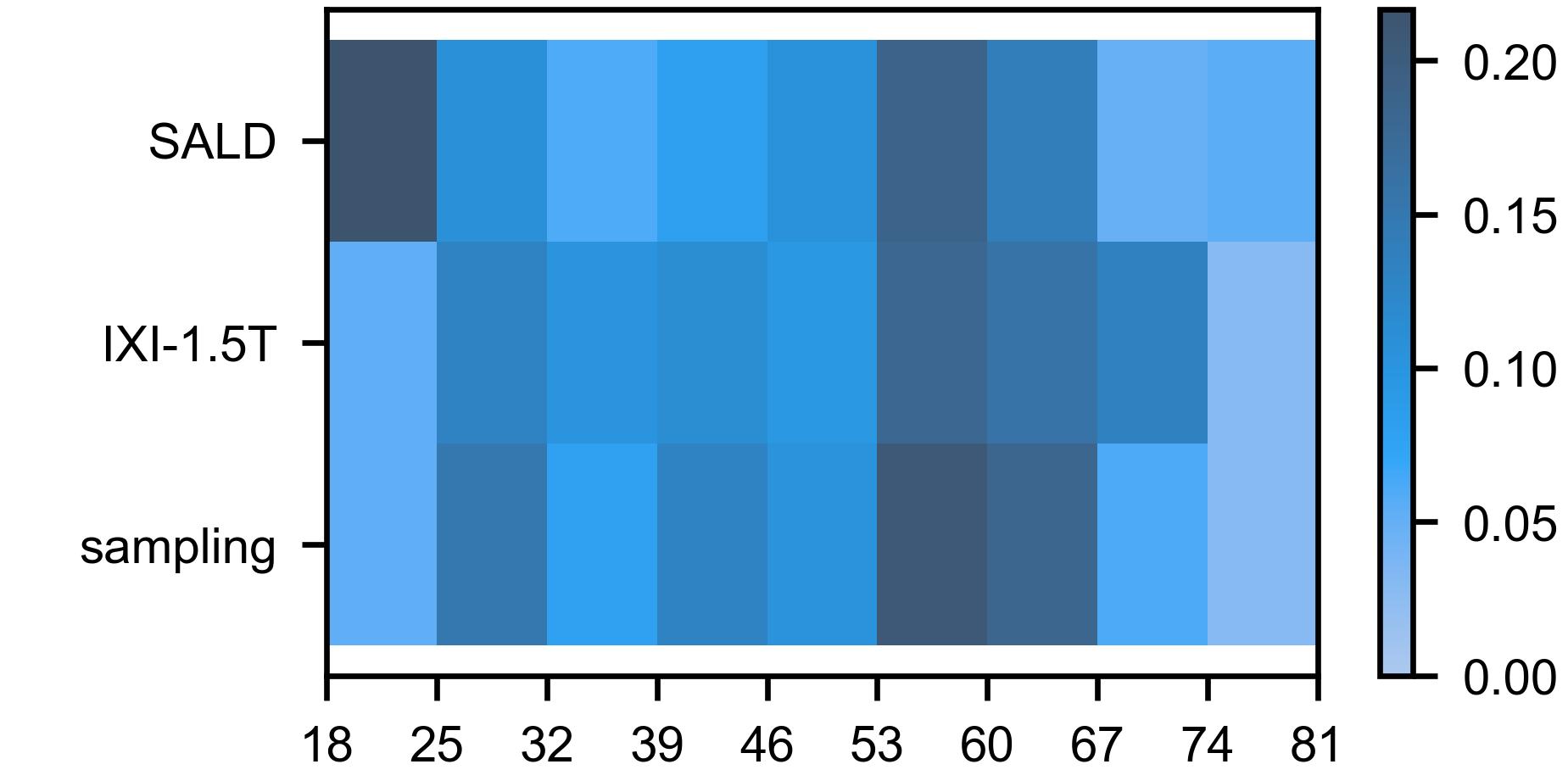}
    \end{subfigure}
    \hfill
    \begin{subfigure}{0.48\linewidth}
        \centering
        \includegraphics[scale=0.9]{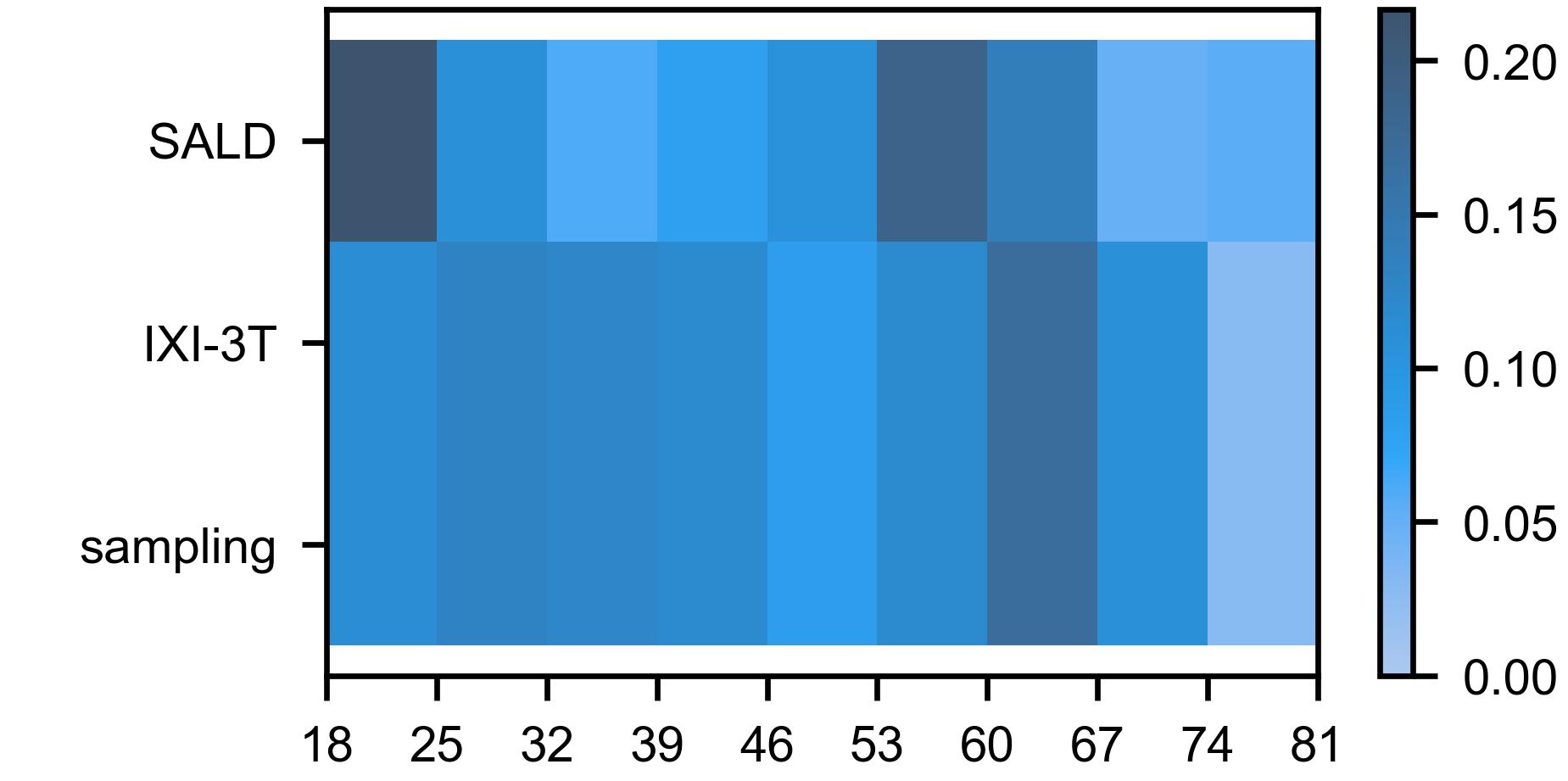}
    \end{subfigure}
    \vspace{1em}
    
    \begin{subfigure}{0.48\linewidth}
        \centering
        \includegraphics[scale=0.9]{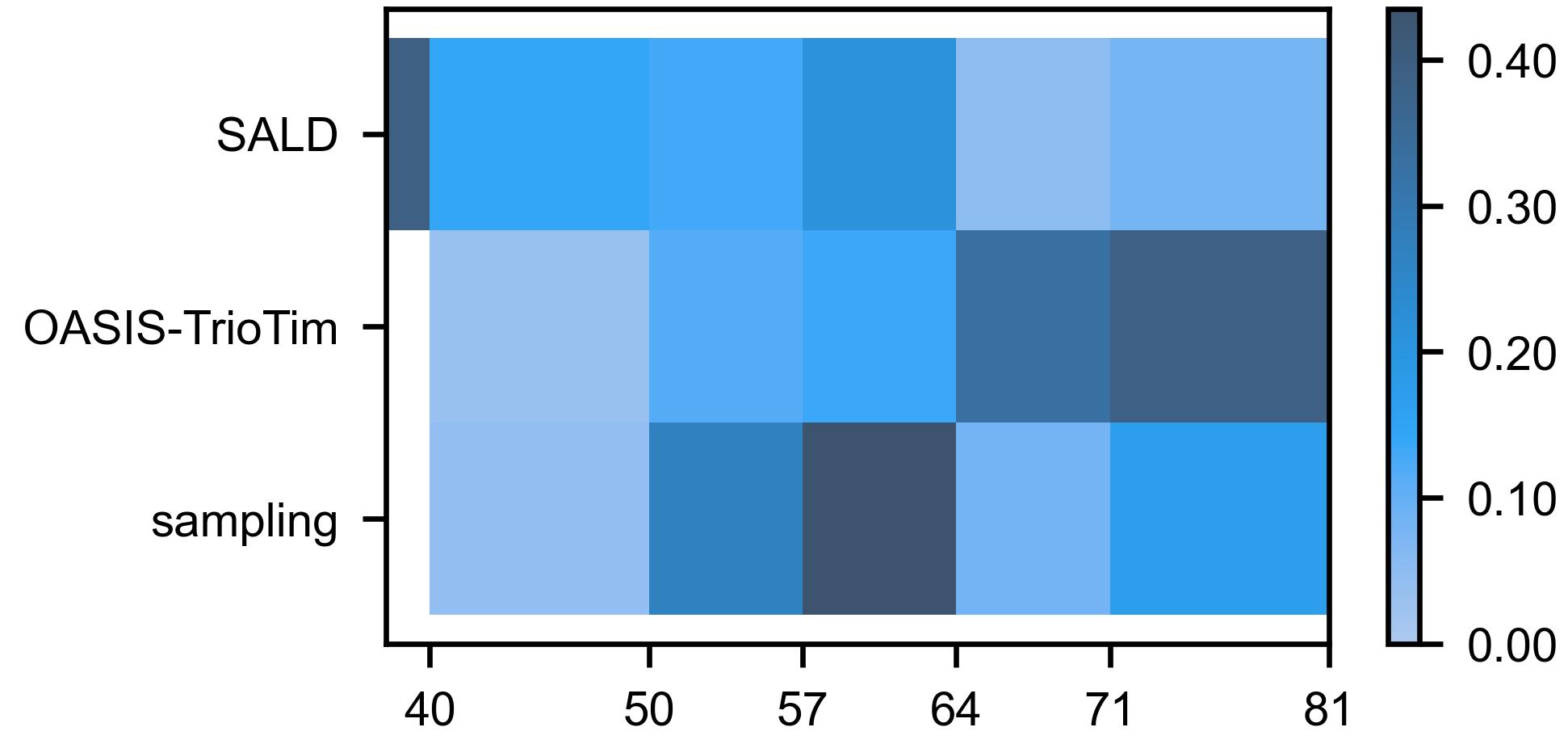}
    \end{subfigure}
    \hfill
    \begin{subfigure}{0.48\linewidth}
        \centering
        \includegraphics[scale=0.9]{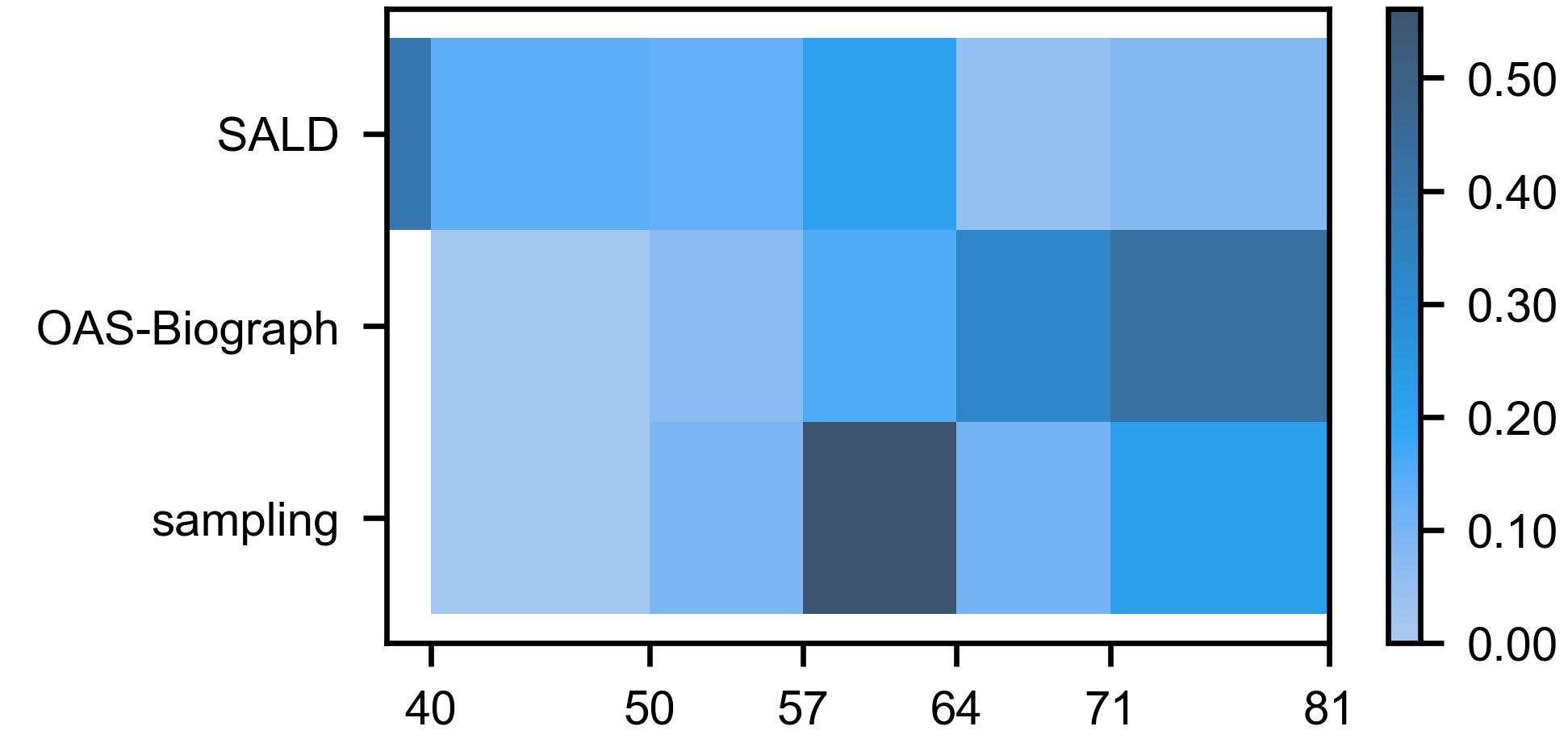}
    \end{subfigure}
    \vspace{1em}

    \begin{subfigure}{0.48\linewidth}
        \centering
        \includegraphics[scale=0.9]{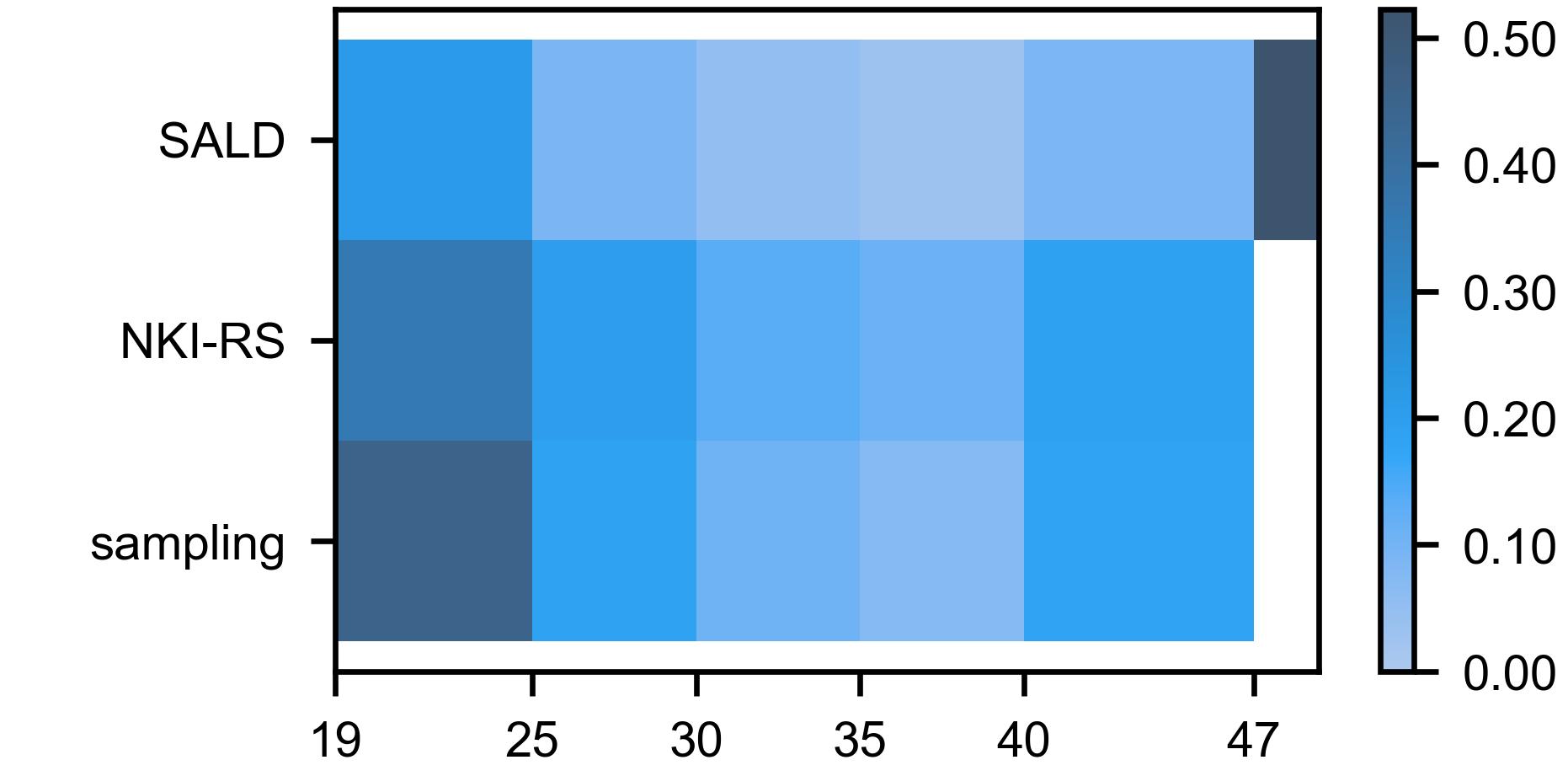}
    \end{subfigure}
    \hfill
    \begin{subfigure}{0.48\linewidth}
        \centering
        \includegraphics[scale=0.9]{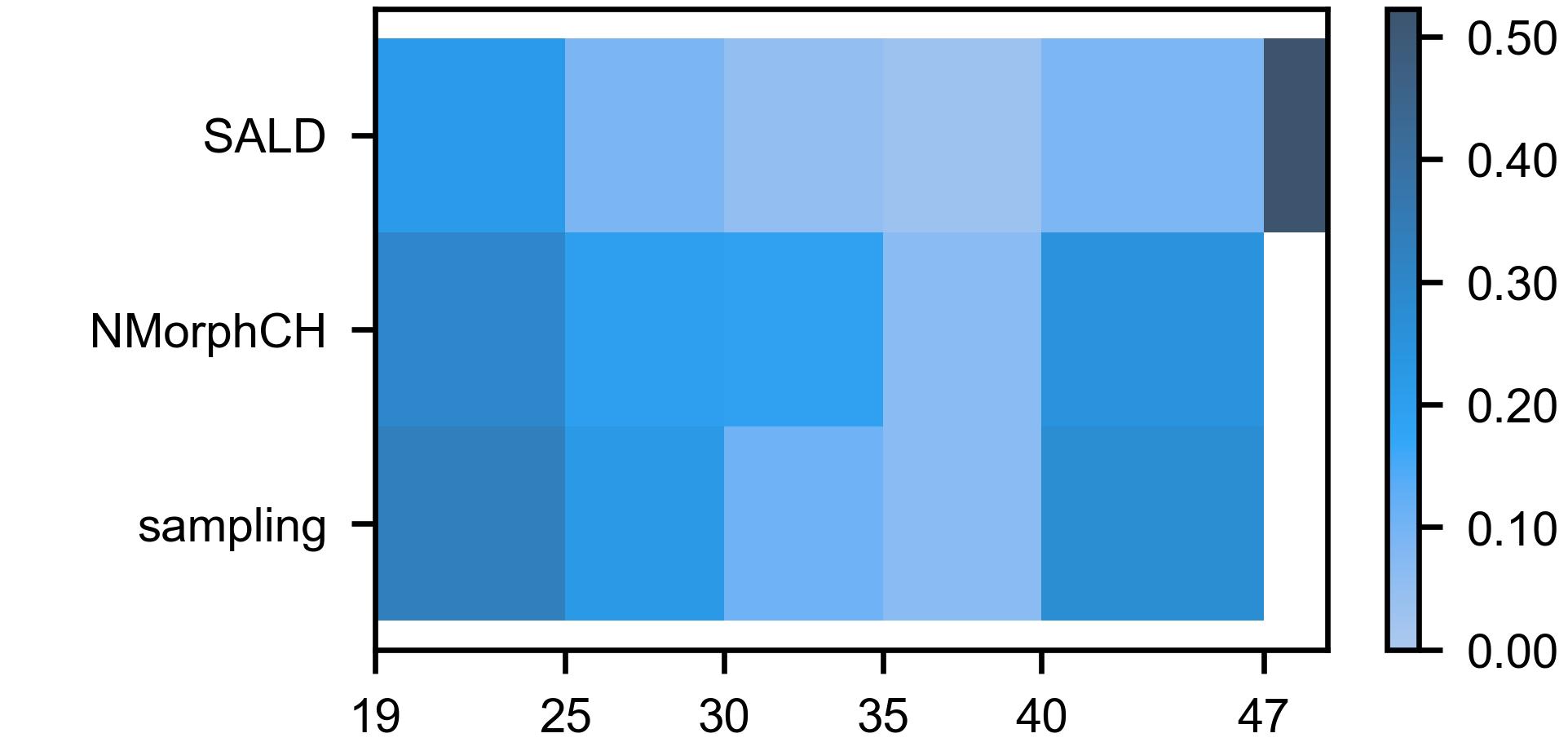}
    \end{subfigure}
    \vspace{1em}

    \begin{subfigure}{0.48\linewidth}
        \centering
        \includegraphics[scale=0.9]{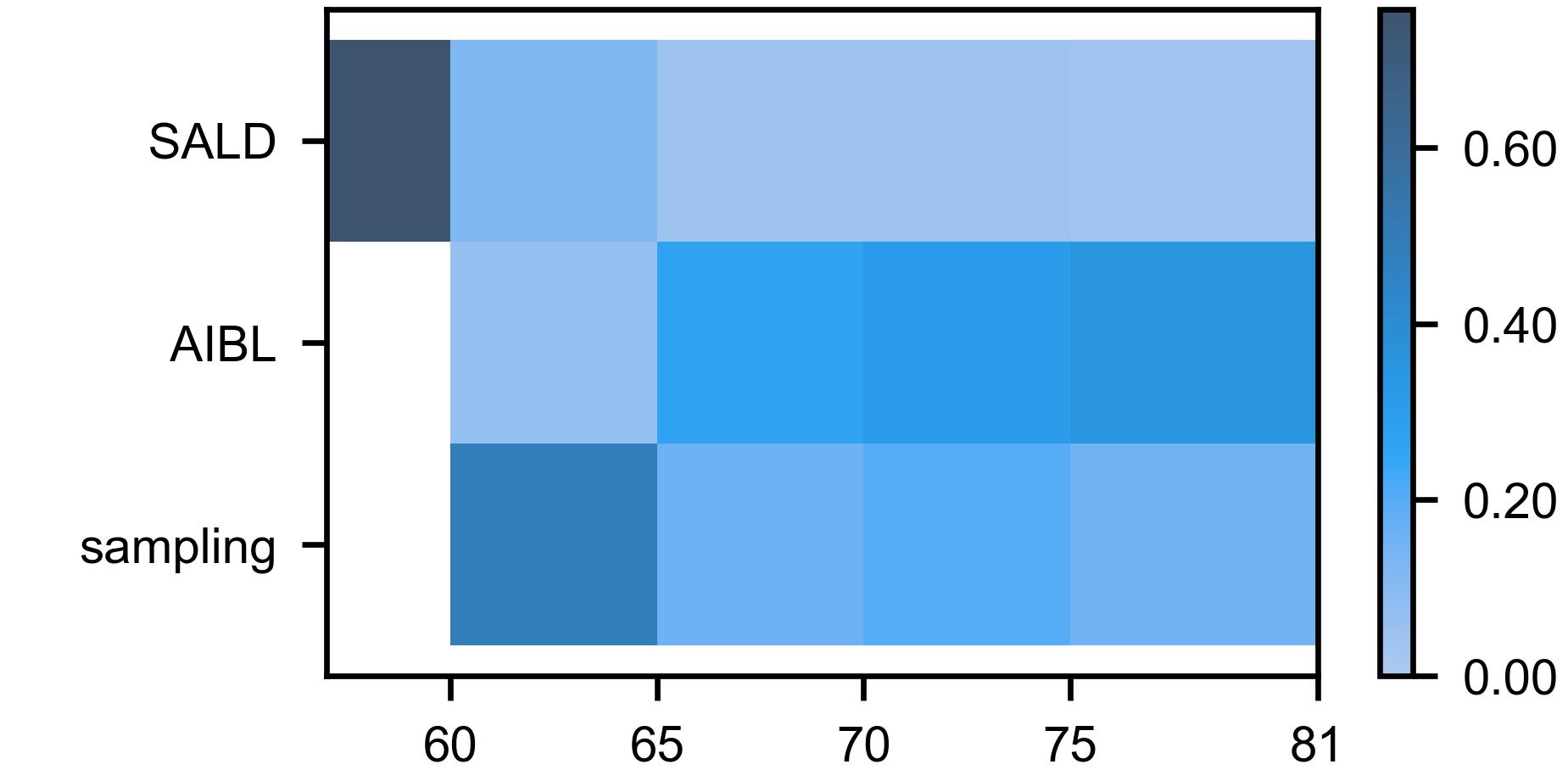}
    \end{subfigure}
    \hfill
    \begin{subfigure}{0.48\linewidth}
        \centering
        \includegraphics[scale=0.9]{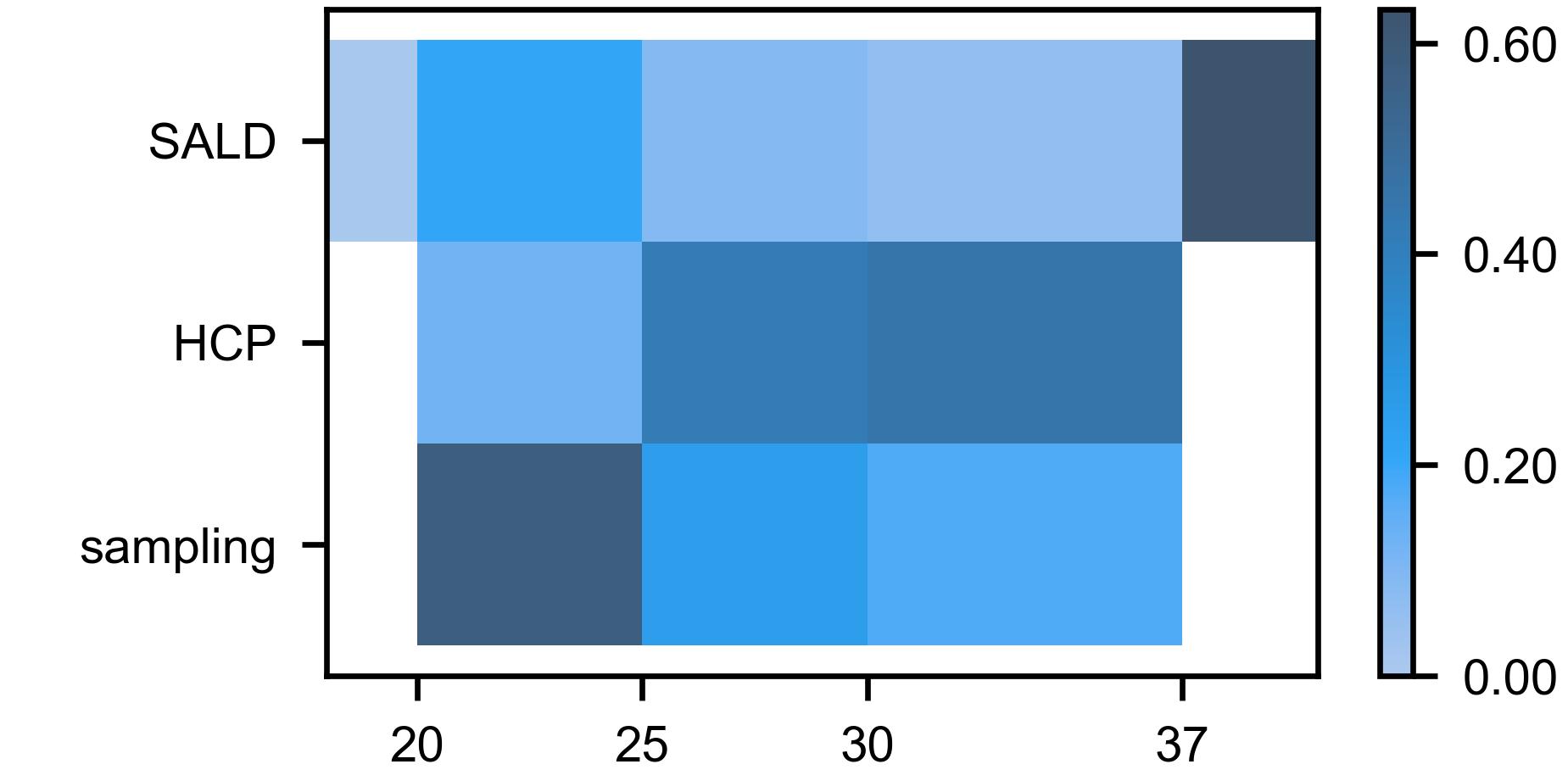}
    \end{subfigure}
    \vspace{1em}

    \begin{subfigure}{0.48\linewidth}
        \centering
        \includegraphics[scale=0.9]{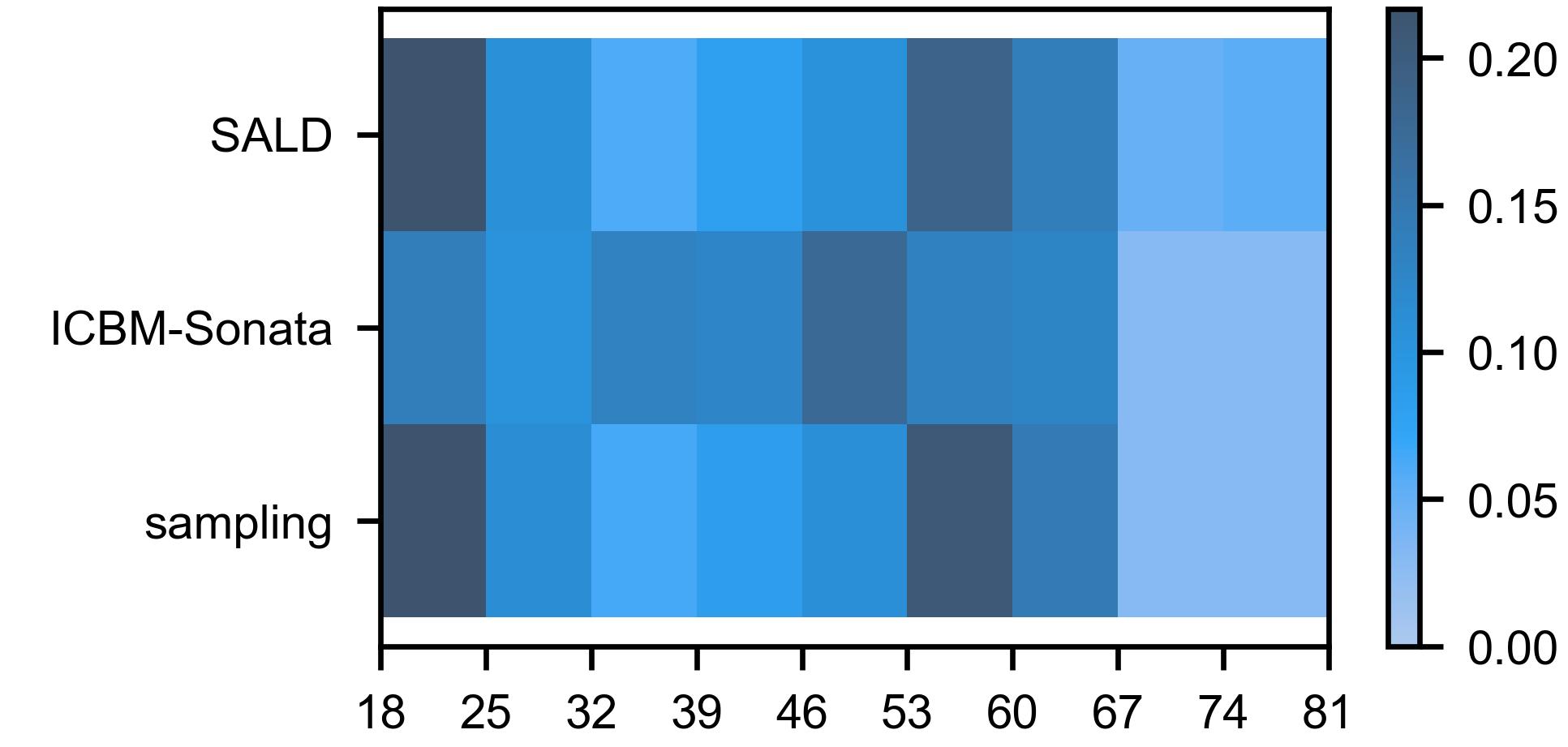}
    \end{subfigure}
    \hfill
    \begin{subfigure}{0.48\linewidth}
        \centering
        \includegraphics[scale=0.9]{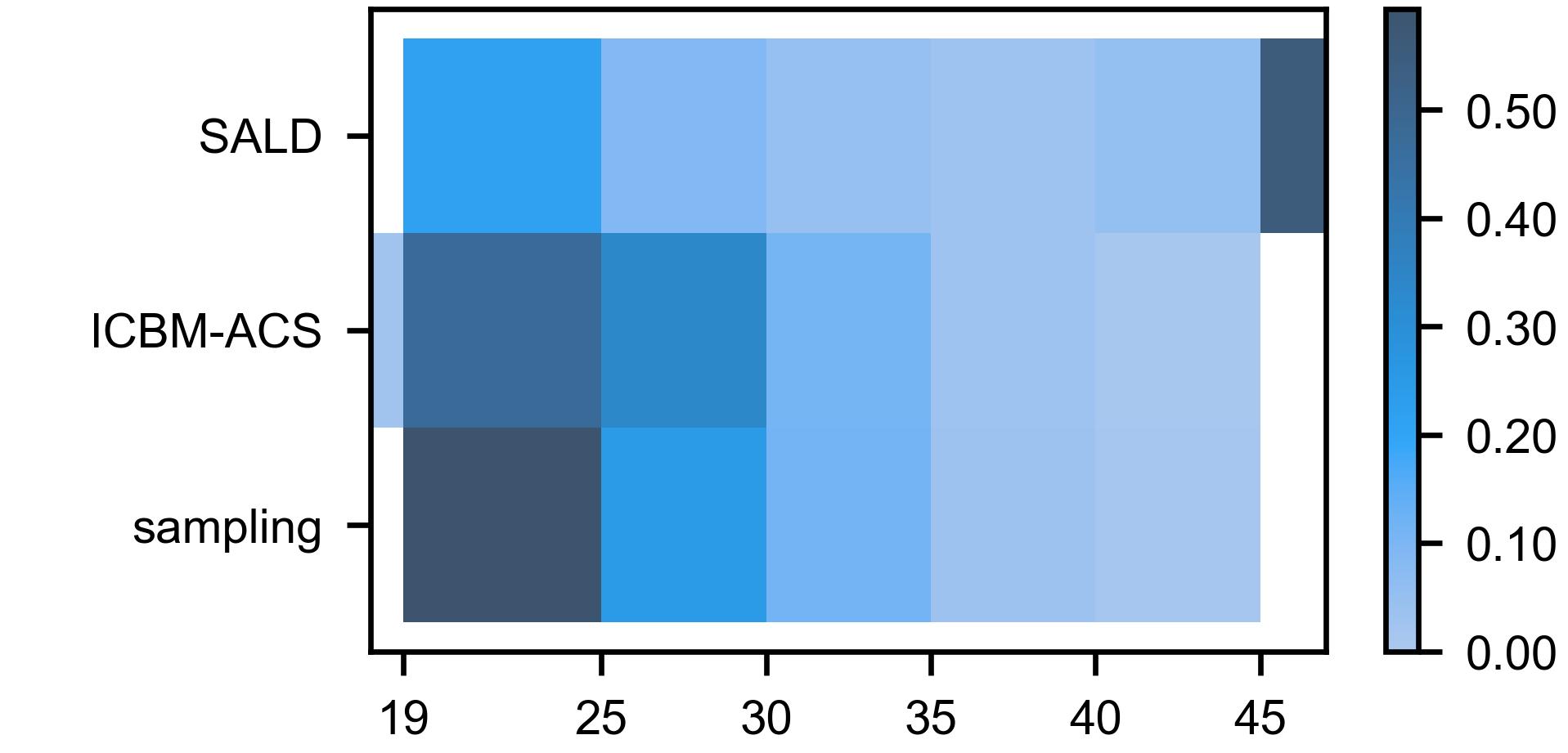}
    \end{subfigure}
    
    \caption{\textbf{Probability distributions of age and sampling for IGUANe training in the Training dataset.} Each subfigure indicates the age distributions of SALD and of a source site as well as the sampling probabilities used for training. The graduations on the x-axes indicate the age ranges used for bias sampling. The color intensities indicate the proportion for each age range.}
    \label{fig:probas_sampling}
\end{figure}

\section{CALAMITI intensity normalization}
\label{appendix_calimiti_norm}
The 3D T1w image provided in the online repository of CALAMITI did not seem to have been WS-normalized in a classical way as it contained no negative values. We therefore determined the normal appearing white-matter mask of the volume - using the code used in CALAMITI\footnotemark[10] - and computed the mean and the standard deviation inside this mask (mean = 1016.9; standard deviation = 7.8). Then, in order to implement the CALAMITI preprocessing, we matched these two values by scaling/shifting the MR intensities (after the three preprocessing steps explained online).

\section{Intensity shifts for structural similarity index}
\label{appendix_ssim_shifts}
To avoid negative values that may hinder the SSIM computations, we added the following constant intensities to the images in our experiment on traveling subjects (section \ref{sec:meth_travSubs}): WS +160; HM +10; CALAMITI +230 before harmonization, +10 after harmonization.

\section{Supplementary details for the predictive models}
\label{appendix_predictive_models}
For both age regression and CN/AD classification, we trained predictive models for 400 epochs with a batch size of 16. We used an Adam optimizer \citep{Kingma2017} with a linear learning rate decay from 0.001 to 0.0001. A data-augmentation consisted in a random translation ($\pm$ 5 voxels) along the three orthogonal axes and a random rotation ($\pm$ 10 °) along a randomly selected orthogonal plane (rotation applied with a probability 1/2).

Each HM-normalized MR image was scaled/shifted with constants so that 1st and 99th percentiles of the brain intensities were -0.5 and 0.5, respectively. Each WS-normalized MR image was scaled/shifted with constants so that the mean and the standard deviation of the normal appearing white-matter was 0.7 and 0.01, respectively.

The code we used in this study is available in our online repository.

\section{Visualization of MR images harmonized with the different methods}
\label{appendix_slices_allMeths}
\setcounter{figure}{0}

\begin{figure}
    \centering
    
    \begin{subfigure}{\linewidth}
        \centering
        \rotatebox[origin=lB]{90}{\footnotesize\parbox{2em}{IGUANe/HM/WS\\preproc}}
        \includegraphics{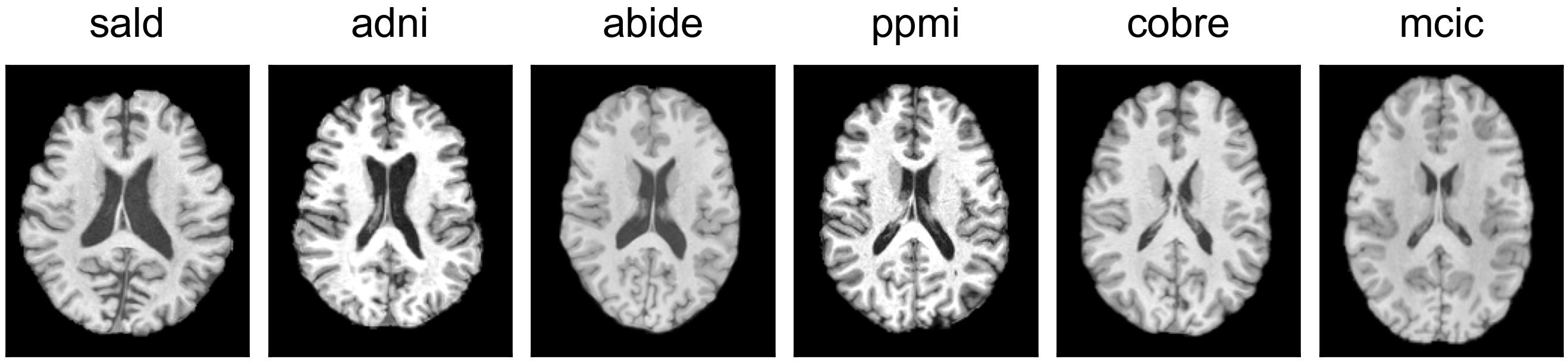}
        \includegraphics{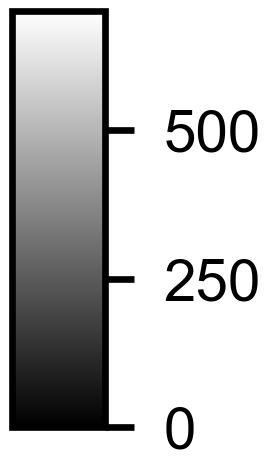}
    \end{subfigure}

    \begin{subfigure}{\linewidth}
        \centering
        \rotatebox[origin=lB]{90}{\footnotesize\parbox{2em}{IGUANe\\harmonized}}
        \includegraphics{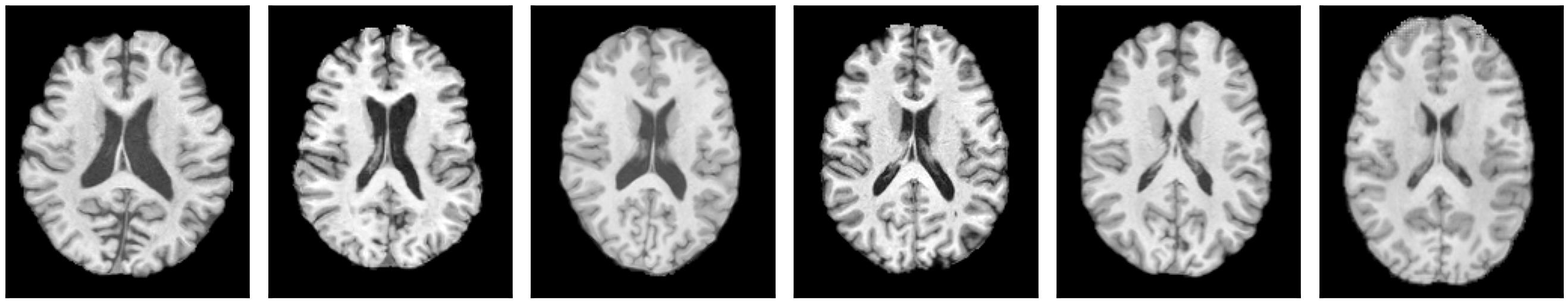}
        \includegraphics{figG1_iguane_cbar.jpg}
    \end{subfigure}

    \begin{subfigure}{\linewidth}
        \centering
        \rotatebox[origin=lB]{90}{\footnotesize\parbox{2em}{HM\\harmonized}}
        \includegraphics{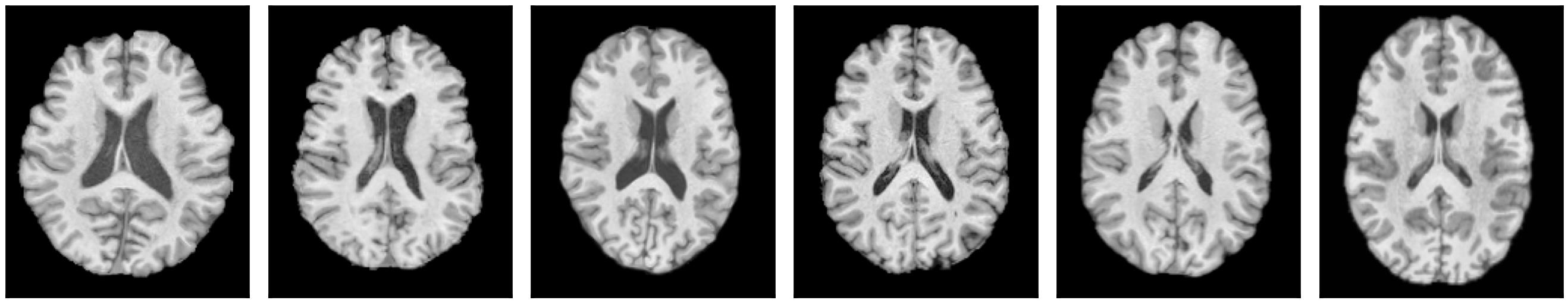}
        \includegraphics{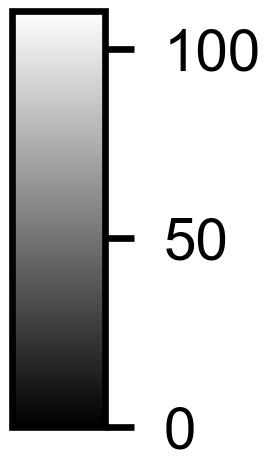}
    \end{subfigure}

    \begin{subfigure}{\linewidth}
        \centering
        \rotatebox[origin=lB]{90}{\footnotesize\parbox{2em}{WS\\harmonized}}
        \includegraphics{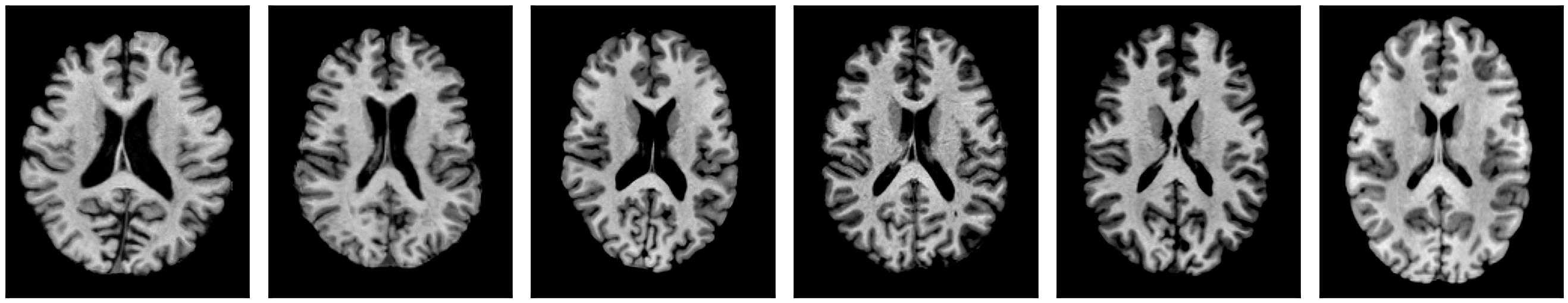}
        \includegraphics{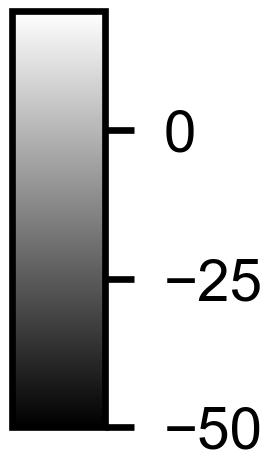}
    \end{subfigure}

    \begin{subfigure}{\linewidth}
        \centering
        \rotatebox[origin=lB]{90}{\footnotesize\parbox{2em}{STGAN\\preproc}}
        \includegraphics{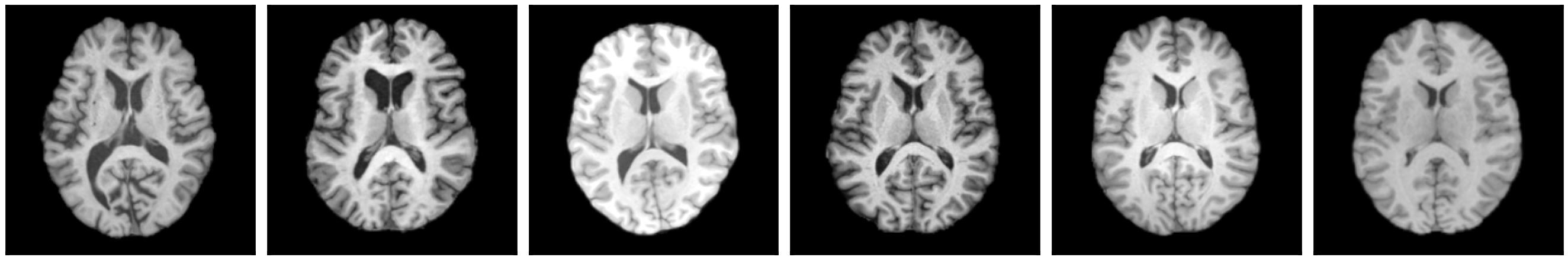}
        \includegraphics{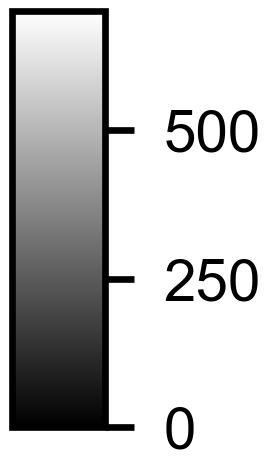}
    \end{subfigure}

    \begin{subfigure}{\linewidth}
        \centering
        \rotatebox[origin=lB]{90}{\footnotesize\parbox{2em}{STGAN\\harmonized}}
        \includegraphics{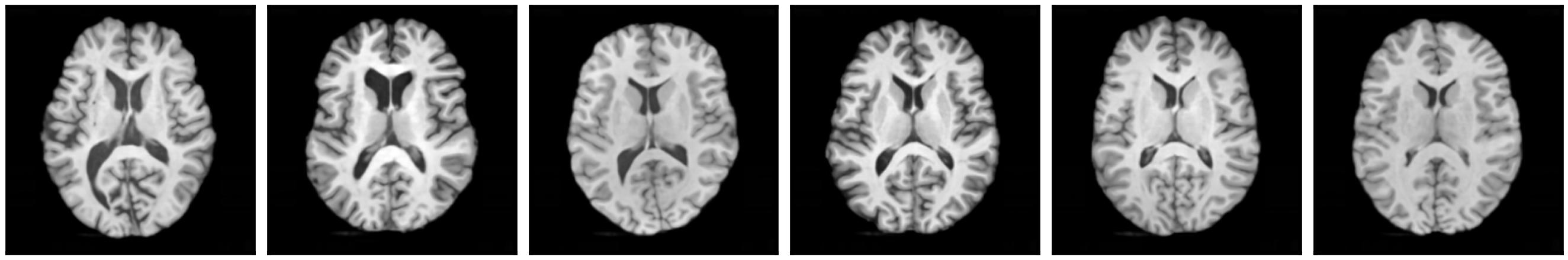}
        \includegraphics{figG1_stgan_cbar.jpg}
    \end{subfigure}

    \begin{subfigure}{\linewidth}
        \centering
        \rotatebox[origin=lB]{90}{\footnotesize\parbox{2em}{CALAMITI\\preproc}}
        \includegraphics{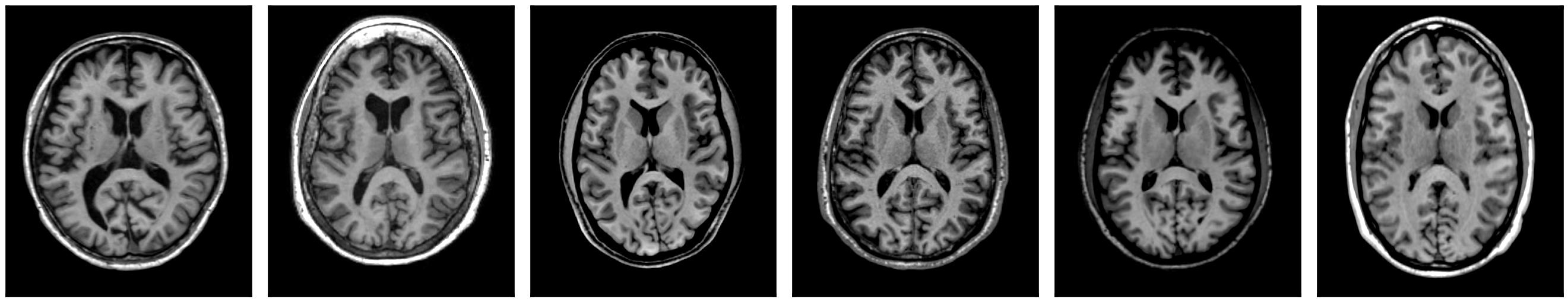}
        \includegraphics{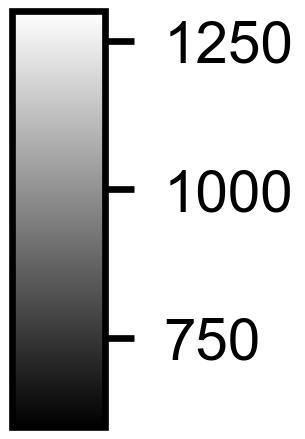}
    \end{subfigure}

    \begin{subfigure}{\linewidth}
        \centering
        \rotatebox[origin=lB]{90}{\footnotesize\parbox{2em}{CALAMITI\\harmonized}}
        \includegraphics{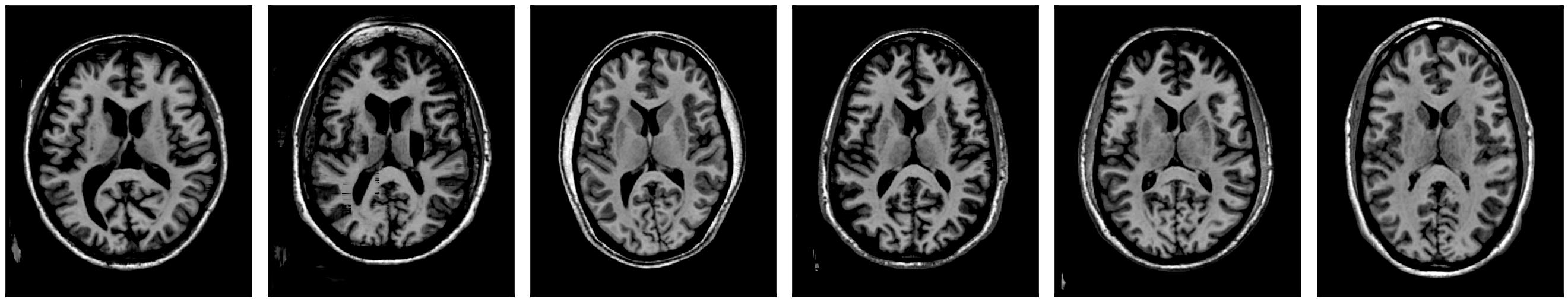}
        \includegraphics{figG1_calam_cbar.jpg}
    \end{subfigure}

    \caption{\textbf{Visualization of MR images harmonized with the different methods implemented.} One image from the reference domain (SALD) and one from each study in the Generalization dataset were randomly sampled, with a middle axial slice shown for each. \textit{preproc} refers to the images obtained after preprocessing for the corresponding harmonization approach.}
    \label{fig:slicesVisuApp}
\end{figure}

Fig. \ref{fig:slicesVisuApp} shows examples of MR images harmonized using the different methods implemented in this study.

\section{Additional results on the SALD dataset}
\label{appendix_results_sald}
\setcounter{figure}{0}
As we used the SALD dataset as the reference domain for IGUANe harmonization, we expected that the model would only make minimal changes to the SALD MR images. To verify this, we randomly selected 10 MR images from the SALD dataset and computed the Euclidean distance for each image pair before and after harmonization with IGUANe. We additionally computed the intraclass correlation coefficient \citep[ICC][]{Bartko1966} to quantify the consistency of the distances before and after harmonization.

\begin{figure}
    \centering
    \includegraphics{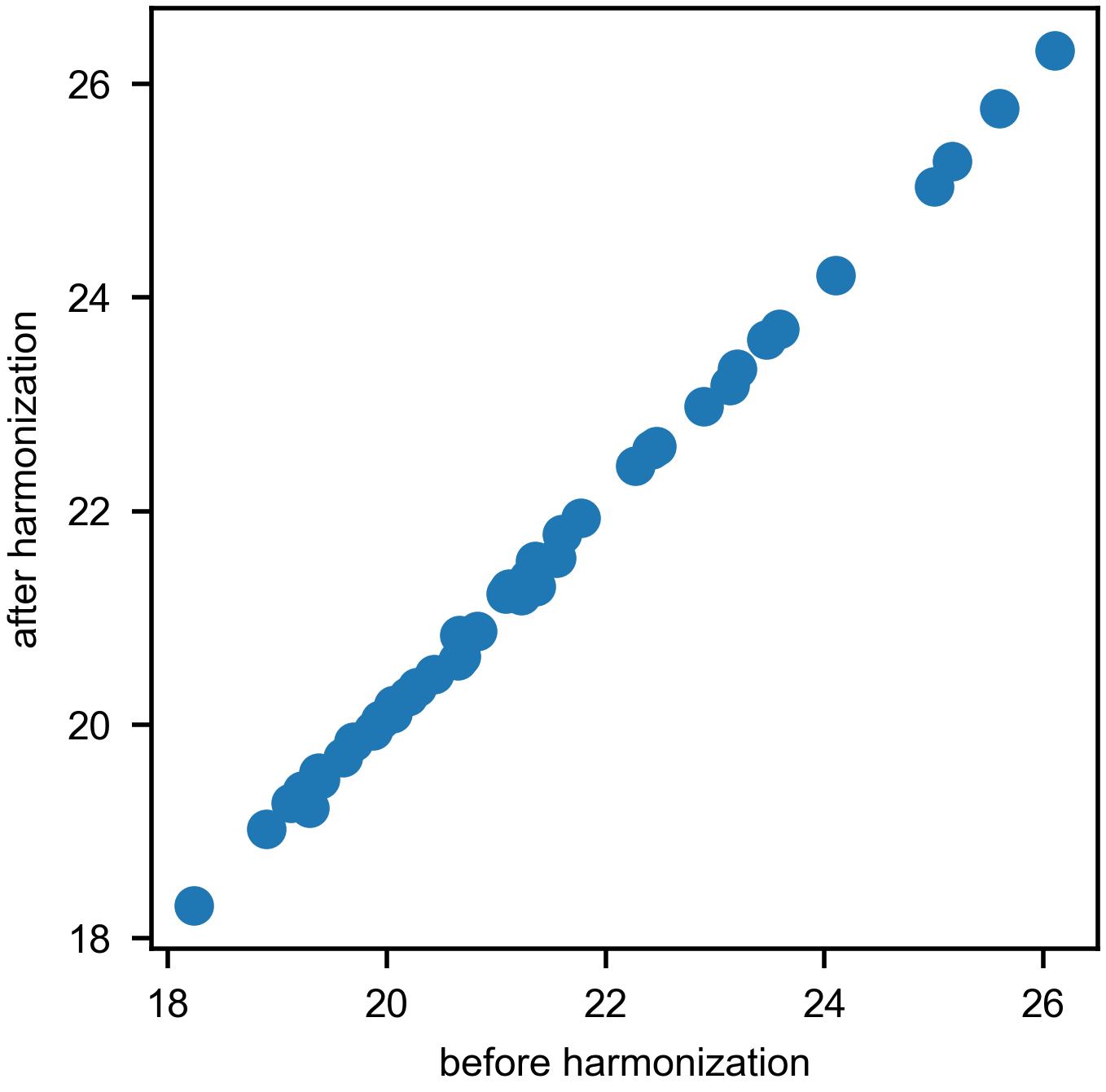}
    \caption{\textbf{Inter-image Euclidean distances in a sample of 10 images from the SALD dataset before and after IGUANe harmonization.} The 45 points correspond to the 45 image pairs. The distances are expressed in tens of thousands.}
    \label{fig:sald_dists}
\end{figure}

The results show an excellent preservation of the distances with IGUANe (Fig. \ref{fig:sald_dists}), which is confirmed by the ICC of 0.998 obtained.

\section{Additional results to study the impact of image resolutions on the correlation between age and gray-matter volume}
\label{appendix_corrGMage_mcic}
\setcounter{table}{0}

It can be seen in Table \ref{tab:resolutions} that most of the images in the Training dataset, except those from HCP, have resolutions near 1mm$^3$. In the Generalization dataset, the MCIC images have higher resolutions in two axes compared to the images from the other studies. To study the impact of this difference in the original resolution, we analyzed the Pearson correlation coefficient between age and GM volume obtained with SPM12 before any preprocessing and after IGUANe harmonization in the Generalization dataset.

The linear anti-correlation was strengthened in ADNI, PPMI, COBRE and ABIDE but weakened in MCIC with IGUANe harmonization (Table \ref{tab:corrGMage_perStudy}), suggesting that the MCIC images were negatively influenced by the registration to 1mm$^3$. However, when gathering the MR images from the six studies to compute a global correlation coefficient, the linear anti-correlation was clearly reinforced (r = -0.759 on the raw images and -0.830 on the harmonized images). The pattern's enhancement was less clear when combining the raw MCIC images with the harmonized images from the other studies (r = -0.803). These results confirm the value of harmonization when analyzing measures from a multicenter dataset.

\begin{table}
    \centering
    \caption{\textbf{Pearson correlation coefficients between age and gray-matter volume per study in the Generalization dataset.}}
    \begin{tabular}{>{\raggedright}m{0.20\linewidth}>{\raggedright}m{0.12\linewidth}>{\raggedright}m{0.12\linewidth}>{\raggedright}m{0.12\linewidth}>{\raggedright}m{0.12\linewidth}>{\raggedright\arraybackslash}m{0.12\linewidth}}
        \hline
        &\textbf{ADNI}&\textbf{MCIC}&\textbf{PPMI}&\textbf{COBRE}&\textbf{ABIDE}\\
        \textbf{raw MR images}&-0.350&-0.831&-0.535&-0.472&-0.475\\
        \textbf{after IGUANe harmonization}&-0.514&-0.741&-0.669&-0.691&-0.602\\
        \hline
    \end{tabular}
    \label{tab:corrGMage_perStudy}
\end{table}

\section{Results of the ablation studies}
\label{appendix_ablation}
\setcounter{figure}{0}

\begin{figure}
    \centering
    \begin{subfigure}{\linewidth}
        \centering
        \includegraphics[scale=1]{fig5_legend.jpg}
    \end{subfigure}
    \par\bigskip

    \begin{subfigure}{0.22\linewidth}
        \rotatebox[origin=center]{90}{before harmonization}

        \begin{minipage}{0.13\linewidth}
            \rotatebox[origin=center]{90}{after harmonization}
        \end{minipage}
        \begin{minipage}{0.82\linewidth}
            \includegraphics[scale=1]{fig5c_harm.jpg}
        \end{minipage}
        \caption{IGUANe}
    \end{subfigure}
    \hspace{1.2em}
    \begin{subfigure}{0.18\linewidth}
        \centering
        \includegraphics[scale=1]{fig5abc_preproc.jpg}
        \includegraphics[scale=1]{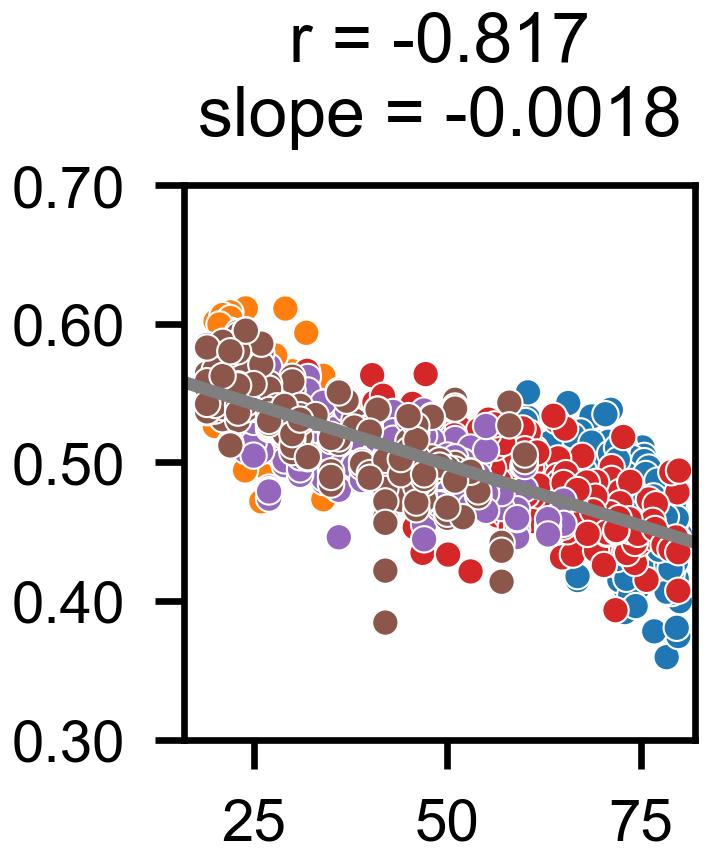}
        \caption{IGUANe\_abl1}
    \end{subfigure}
    \hspace{1.2em}
    \begin{subfigure}{0.18\linewidth}
        \centering
        \includegraphics[scale=1]{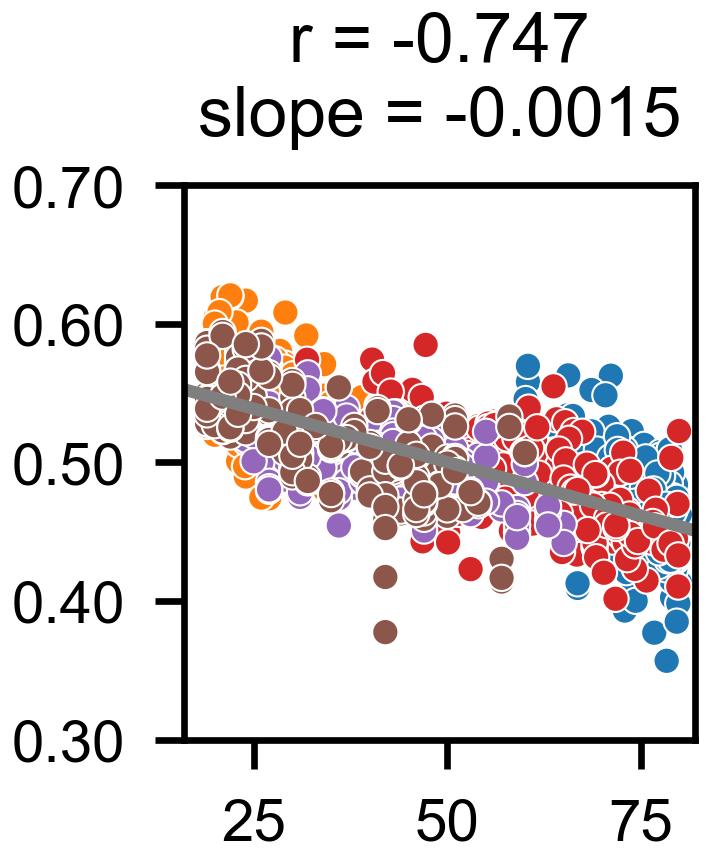}
        \caption{IGUANe\_abl2}
    \end{subfigure}
    \caption{\textbf{Correlation between age and gray-matter (GM) volume in the Generalization dataset for the ablation studies.} The X and Y axes correspond to ages and GM volumes (divided by the total intracranial volume), respectively. The linear least-squares regression line is plotted on each subfigure.}
    \label{fig:corr_gmAge_abl}
\end{figure}

\begin{figure}
    \centering
    \includegraphics{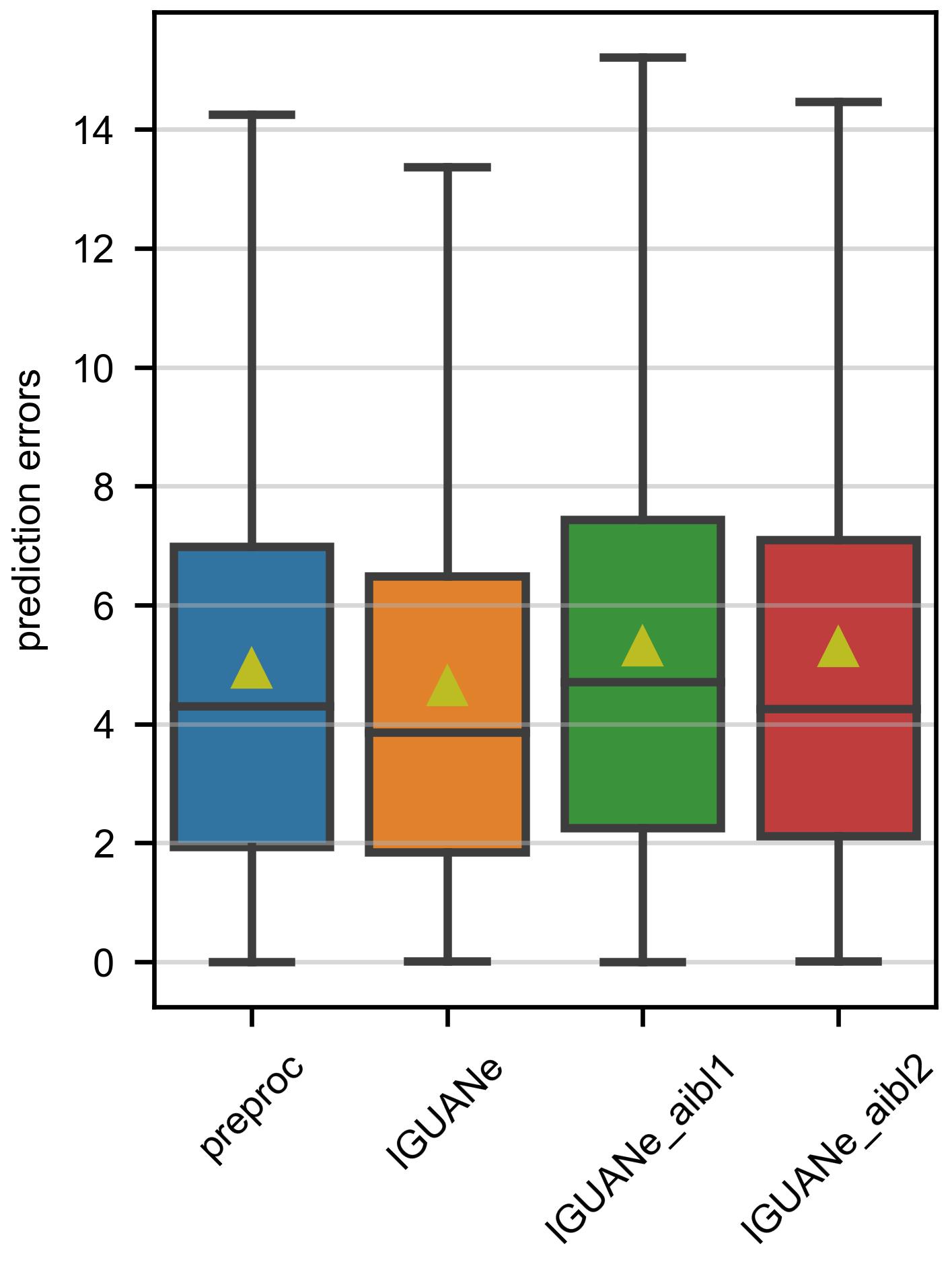}
    \caption{\textbf{Age prediction errors in the Generalization dataset for the ablation studies.} \textit{preproc} refers to the images obtained after the IGUANe preprocessing. The triangles indicate the means.}
    \label{fig:brainAge_abl}
\end{figure}

Results of the ablation studies are presented in Fig. \ref{fig:corr_gmAge_abl} and Fig. \ref{fig:brainAge_abl}. In these two figures, IGUANe\_abl1 is the version in which the $N$ forward discriminators were replaced by a single one. IGUANe\_abl2 is the same but without the age-based sampling strategy.

The analysis of GM volumes in relation to age revealed that the strongest linear anti-correlation and steepest regression slope were achieved with IGUANe (Fig. \ref{fig:corr_gmAge_abl}). In terms of age prediction, the errors were higher with the ablated versions of IGUANe (Fig. \ref{fig:brainAge_abl}).

\section{Additional results on the comparison of hippocampal volumes}
\label{appendix_hippocamp}
\setcounter{figure}{0}
We reproduced the original experiment (section \ref{sec:meth_hippoVols}) by selecting MR images acquired with the Siemens Verio scanner from the AIBL study. Specifically, we selected 71 MR images from 69 participants diagnosed as either CN or AD to ensure similar age distributions (the average age was 75.14 and 75.57 for the CN and AD group, respectively). This approach aimed to focus the analysis on a single scanner to isolate the between-group effects. The results are shown in Fig. \ref{fig:hippoVols_monoScanner}. The effect sizes were slightly more important than those obtained with a multicenter dataset (Fig. \ref{fig:hippo_vols}), but the impact of each harmonization method was similar.

\begin{figure}
    \centering
    \includegraphics{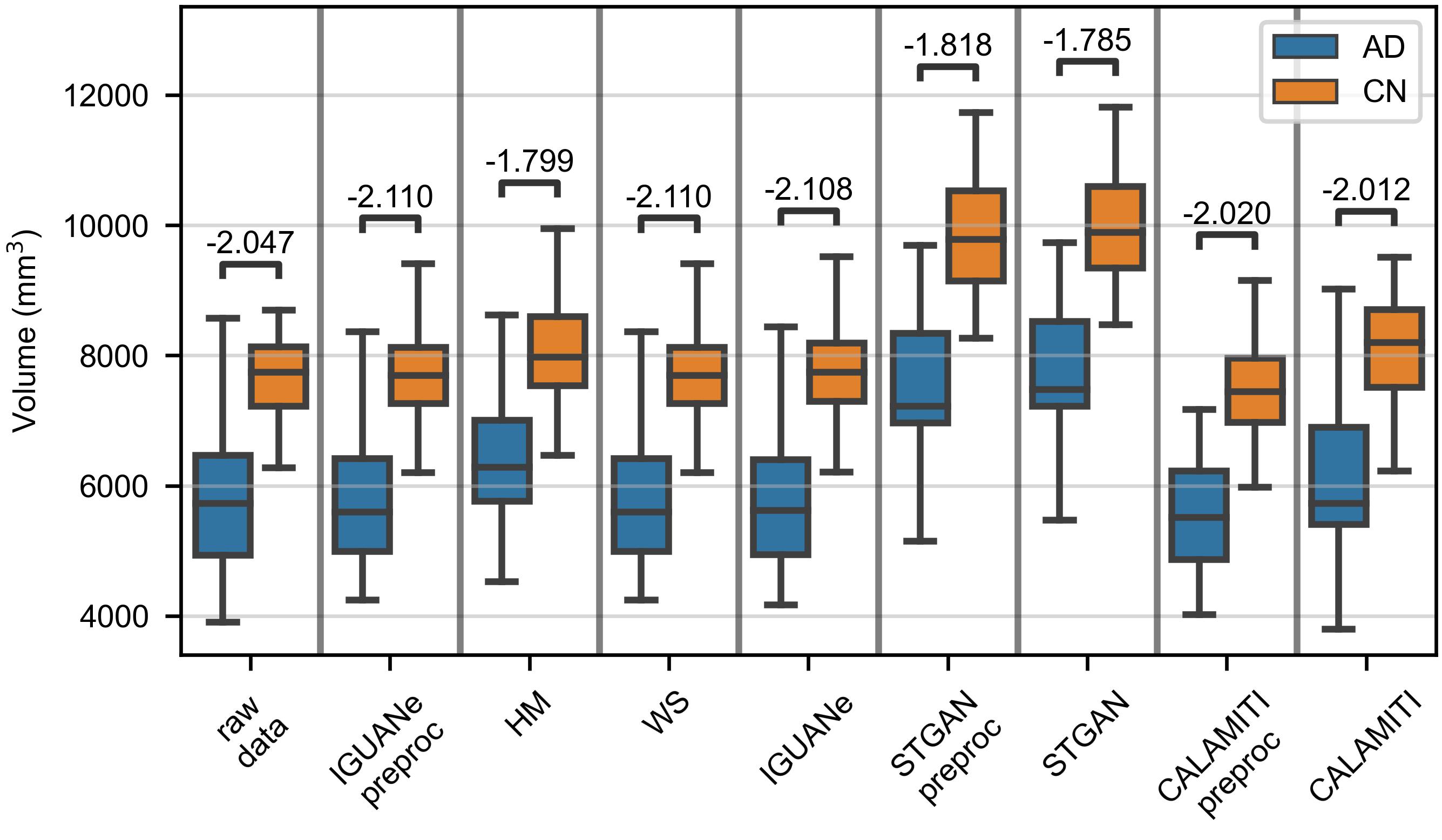}
    \caption{\textbf{Hippocampal volumes and comparison between healthy participants (CN) and subjects with Alzheimer's disease (AD) in the single-scanner dataset from AIBL.} \textit{preproc} refers to the images obtained after preprocessing for the corresponding harmonization approach. Cohen’s d scores comparing the CN and AD groups are above the boxplots.}
    \label{fig:hippoVols_monoScanner}
\end{figure}

\begin{figure}
    \centering
    \includegraphics{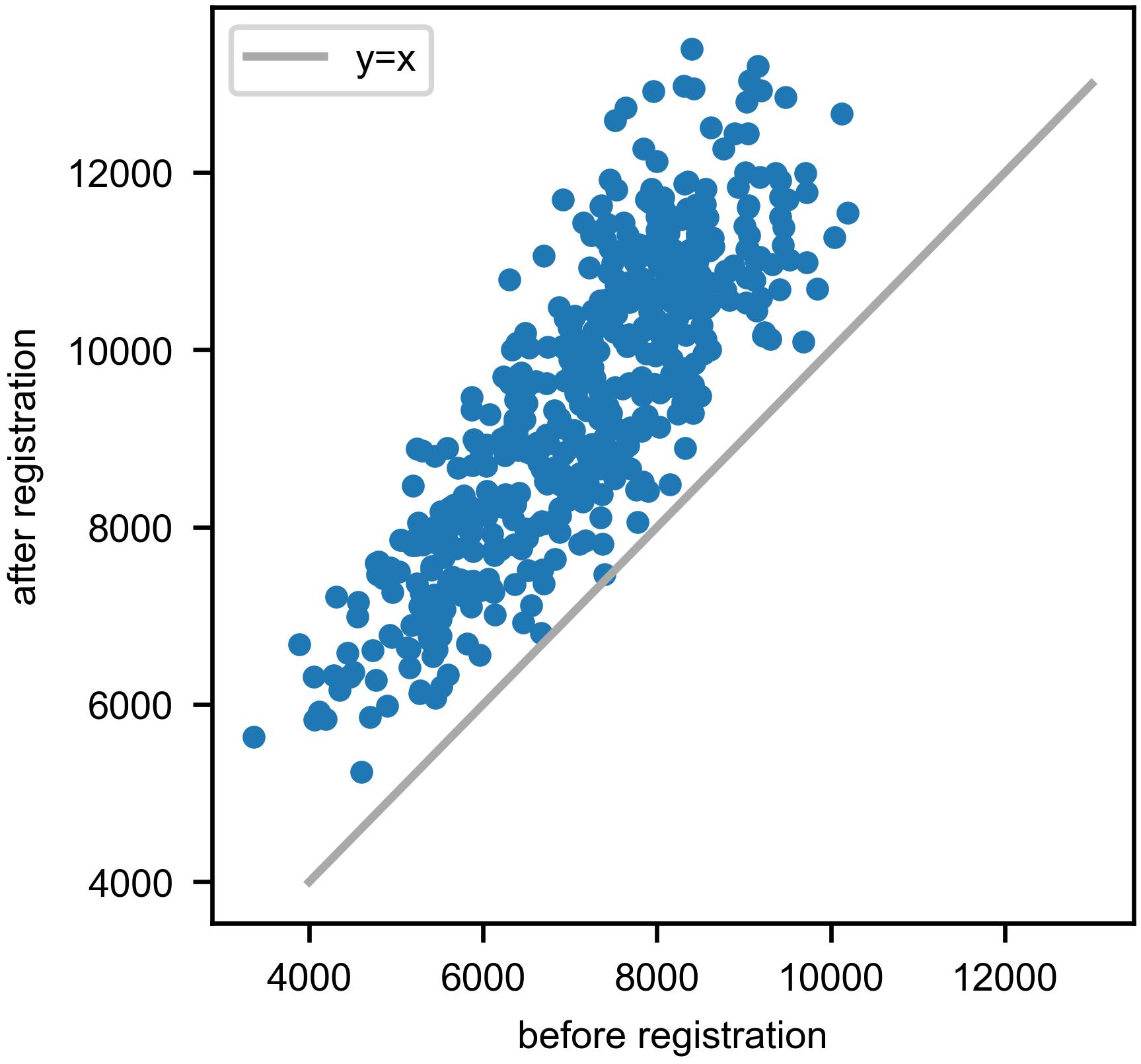}
    \caption{\textbf{Hippocampal volumes before and after registration in the STGAN preprocessing in the Clinical dataset.} The volumes are expressed in mm$^3$.}
    \label{fig:reg_effects}
\end{figure}

We also conducted complementary analyses to understand the origin of the over-estimation of the hippocampal volume in the STGAN preprocessing (Fig. \ref{fig:hippo_vols} and \ref{fig:hippoVols_monoScanner}). To this end, we computed the volumes with SynthSeg on the 500 MR images sampled from the Clinical dataset (section \ref{sec:meth_hippoVols}) after each preprocessing step and found that the linear registration with nine degrees of freedom was the origin of the over-estimation (Fig. \ref{fig:reg_effects}).

\bibliographystyle{elsarticle-harv}\biboptions{authoryear}
\bibliography{refs}

\footnotetext[3]{\url{https://brain-development.org/ixi-dataset/} accessed 2022-01-15}
\footnotetext[4]{\url{http://otto.fsm.northwestern.edu/} accessed 2020-01-15}
\footnotetext[5]{\url{https://www.humanconnectome.org/study/hcp-young-adult} accessed 2020-01-15}
\footnotetext[6]{\url{https://ida.loni.usc.edu/} accessed 2023-03-12}
\footnotetext[7]{\url{https://adni.loni.usc.edu/} accessed 2023-03-12}
\footnotetext[8]{\url{https://www.ppmi-info.org/} accessed 2023-06-01}
\footnotetext[9]{\url{https://fcon_1000.projects.nitrc.org/indi/abide/} accessed 2023-03-12}
\footnotetext[10]{\url{https://github.com/USCLoBeS/style_transfer_harmonization} accessed 2023-07-15}
\footnotetext[11]{\url{https://iacl.ece.jhu.edu/index.php?title=CALAMITI} accessed 2023-07-27}

\end{document}